\pdfoutput=1
\documentclass{article} %
\usepackage{iclr2022_conference,times}

\usepackage{amsmath,amsfonts,bm}

\def\eqref#1{equation~\ref{#1}}

\def\1{\bm{1}}

\DeclareMathAlphabet{\mathsfit}{\encodingdefault}{\sfdefault}{m}{sl}
\SetMathAlphabet{\mathsfit}{bold}{\encodingdefault}{\sfdefault}{bx}{n}

\usepackage{hyperref}
\usepackage{url}

\usepackage{graphicx}
\usepackage{booktabs}
\usepackage{subcaption}
\usepackage{multirow}
\usepackage{xcolor}

\title{Train Short, Test Long: Attention with Linear Biases Enables Input Length Extrapolation }

\author{Ofir Press$^{1,2}$  \quad Noah A. Smith$^{1,3}$ \quad Mike Lewis$^2$ \\
\\ $^1$Paul G. Allen School of Computer Science \& Engineering, University of Washington
\\ $^2$Facebook AI Research
\\ $^3$Allen Institute for AI
\\ \texttt{ofirp@cs.washington.edu}}

\newcommand{\nascomment}[1]{{}}

\usepackage{xspace}
\newcommand{\li}{\ensuremath{L_{\textit{valid}}}\xspace}
\newcommand{\lt}{\ensuremath{L}\xspace}
\newcommand{\al}{ALiBi\xspace}

\iclrfinalcopy 

\begin{document}

\maketitle

\begin{abstract}

Since the introduction of the transformer model by \citet{vaswani}, a fundamental question has yet to be answered:  how does a model achieve extrapolation at inference time for sequences that are longer than it saw during training? 
We first show that extrapolation can be enabled by simply changing the position representation method, though we find that current methods do not allow for \emph{efficient} extrapolation. 
We therefore introduce a simpler and more efficient position method, Attention with Linear Biases (ALiBi). ALiBi does not add positional embeddings to word embeddings;  instead, it biases query-key attention scores with a penalty that is proportional to their distance. We show that this method trains a 1.3 billion parameter model on input sequences of length 1024 that extrapolates to input sequences of length 2048, achieving the same perplexity as a sinusoidal position embedding model trained on inputs of length 2048 but training 11\% faster and using 11\% less memory. 
ALiBi's inductive bias towards recency also leads it to outperform multiple strong position methods on the WikiText-103 benchmark.%
\footnote{Code \& models: \url{https://github.com/ofirpress/attention_with_linear_biases}}

\end{abstract}

\section{Introduction}
\label{sec:intro}

When constructing a transformer-based language model, a major design decision is the length of training sequences, denoted $L$ herein, which has to date  been equivalent to the length of inference sequences.  More context, achieved by larger $L$, improves predictions at inference time.  But longer sequences are more expensive to train on.\footnote{Figure~\ref{fig:wt103_train_speeds_all} in the appendix plots training speed, in words per second, against $L$.}

\begin{figure}[h]
\centering
\begin{subfigure}{.5\textwidth}
  \centering
  \includegraphics[width=\linewidth]{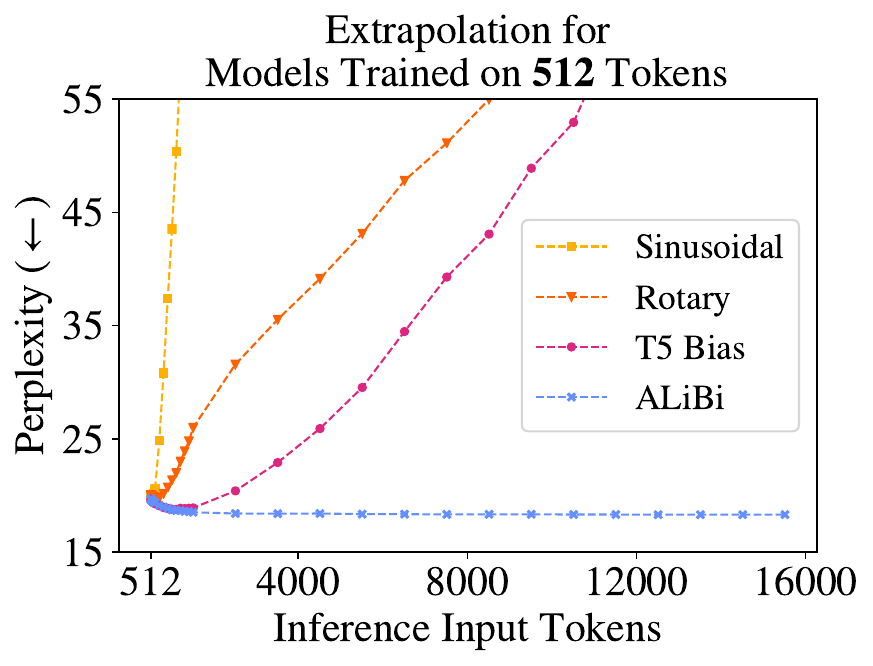}
\end{subfigure}%
\begin{subfigure}{.5\textwidth}
  \centering
  \includegraphics[width=\linewidth]{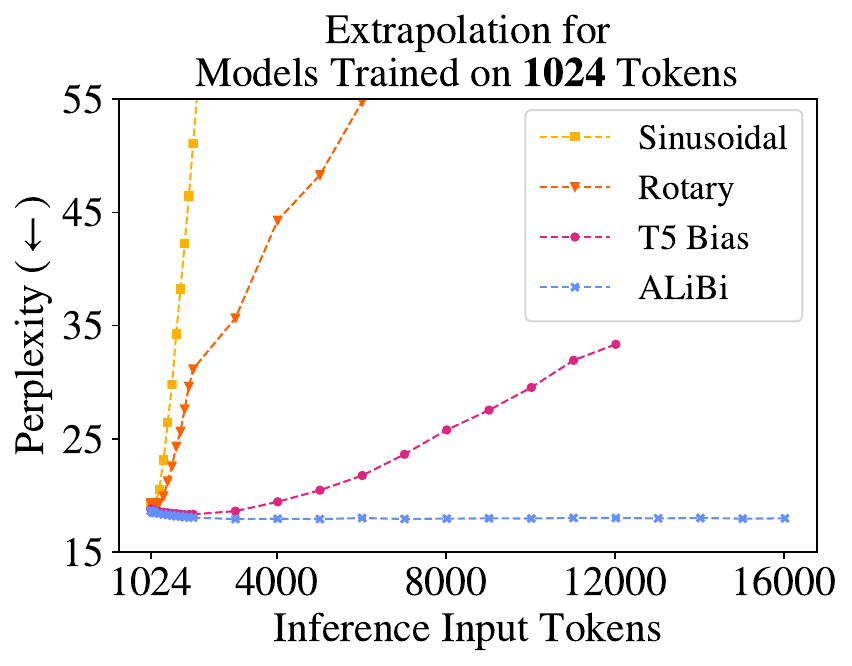}
\end{subfigure}
\caption{Extrapolation:
as the (validation-set's) input sequence gets longer ($x$-axis), current position methods (sinusoidal, rotary, and T5) show degraded perplexity ($y$-axis, lower is better), but our method (\S\ref{sec:ourmethod}) does not.  Models were trained on WikiText-103 with sequences of $L=$ 512 (left) or $L=$ 1,024 (right) tokens.  T5 ran out of memory on our 32GB GPU. For more detail on exact perplexities and runtimes, see Tables~\ref{tab:appendix:wt103_extra_512} and~\ref{tab:appendix:wt103_extra_1024} in the appendix.}

\label{fig:wt103_extra}
\end{figure}

Before transformers, RNN language models were trained on shorter-$L$ sequences and assumed to generalize to longer contexts at inference time~\citep{Mikolov2010RecurrentNN, Mikolov2012ContextDR, zaremba2014recurrent}.  \citet{vaswani}, introducing the transformer, speculated that it ``may [...] extrapolate to sequence lengths longer than the ones encountered during training.''  
We define \emph{extrapolation} as a model's ability to continue performing well as the number of input tokens during validation increases beyond the number of tokens on which the the model was trained. 
We find that transformer language models (LMs)  that use sinusoidal position embeddings have very weak  %
extrapolation abilities; see Figure~\ref{fig:wt103_extra}.

We demonstrate that this failure to extrapolate is caused by the position embedding method.  As shown in Figure~\ref{fig:wt103_extra}, recent alternatives to the original sinusoidal position method \citep{roformer,t5} have improved extrapolation.  However, the better of these, the T5 bias, is considerably slower than the sinusoidal approach and uses extra memory and parameters (Figure~\ref{fig:wt103_speed_mem}).

We therefore introduce Attention with Linear Biases (ALiBi) to facilitate  efficient extrapolation. %
ALiBi negatively biases attention scores with a linearly decreasing penalty proportional to the distance between the relevant key and query. Our simple approach eliminates position embeddings. %

Compared to a sinusoidal model trained on the same input length, our method requires no additional runtime or parameters and incurs a negligible (0--0.7\%) memory increase.  ALiBi can be implemented by changing only a few lines of existing transformer code.

Using ALiBi, a transformer LM can be trained on short-$L$ sequences and therefore at much lower cost, and it can still be reliably applied to long sequences at runtime. For example, a 1.3 billion parameter LM trained on $L=$ 1024 tokens with ALiBi achieves the same perplexity as a sinusoidal model trained on $L=$ 2048 when both are tested on sequences of 2048 tokens, even though \textit{our model is 11\% faster and uses 11\% less memory. }%

Though performance peaks at around two times the number of tokens that the model was trained on, ALiBi maintains strong performance even on sequences of length 10,000. 
In recently explored settings where NLP training examples are given as context to an LM \citep{gpt3}, our approach will allow exposure to more examples. Additionally, it enables generation of longer outputs.

\section{Current Approaches Do Not Extrapolate Efficiently }
\label{sec:act2}

\begin{figure}
\centering
\begin{subfigure}{.28\textwidth}
  \centering
  \includegraphics[width=\linewidth]{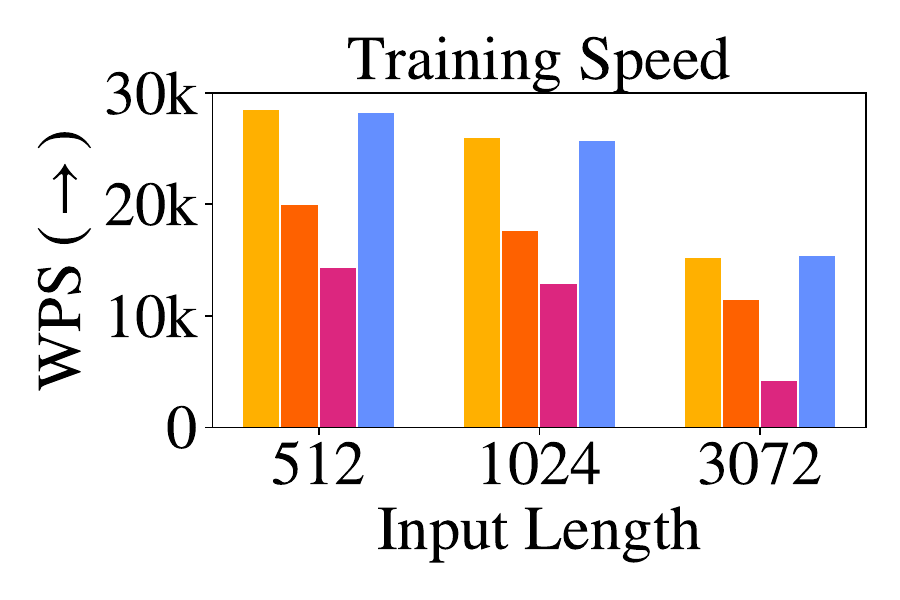}
\end{subfigure}%
\begin{subfigure}{.28\textwidth}
  \centering
  \includegraphics[width=\linewidth]{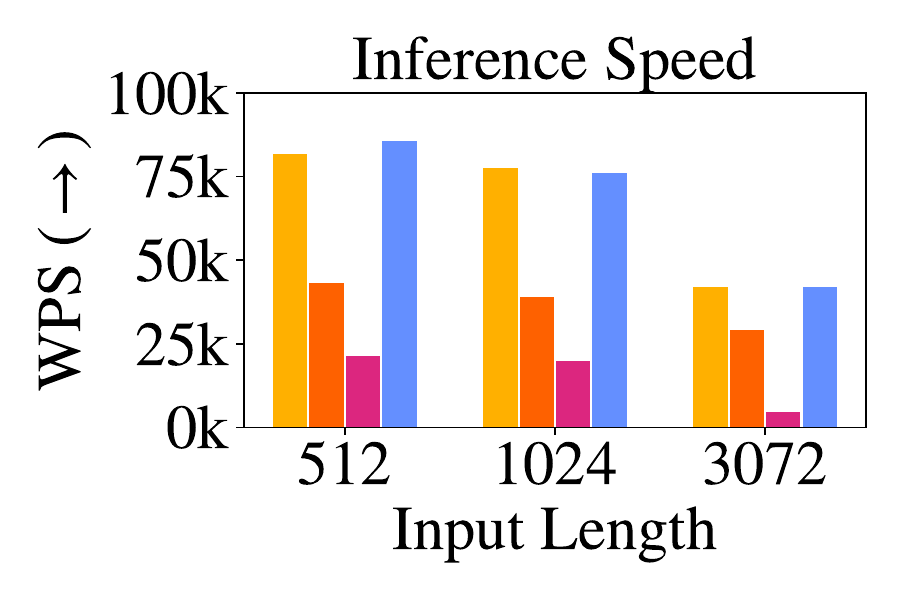}
\end{subfigure}
\begin{subfigure}{.28\textwidth}
  \centering
  \includegraphics[width=\linewidth]{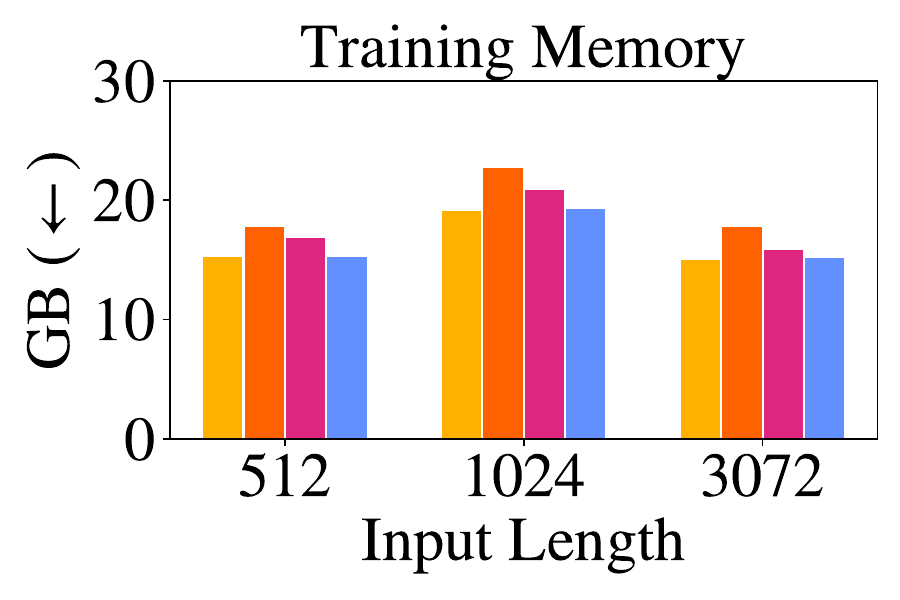}
\end{subfigure}
\begin{subfigure}[]{.14\textwidth}
\centering
\raisebox{3mm}{\includegraphics[width=\linewidth]{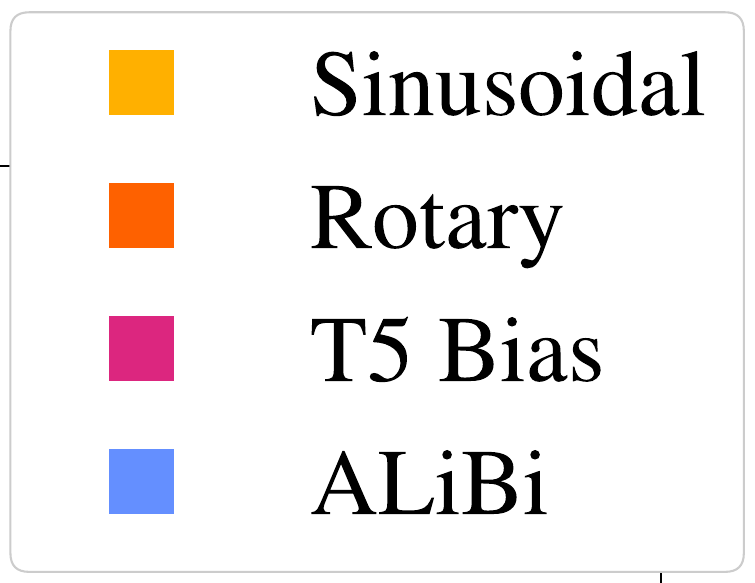}}
\end{subfigure}
\caption{A comparison of batched training, inference speed and memory use of the sinusoidal, rotary, T5 bias, and our ALiBi position methods. The speed differences between our method and the sinusoidal are within 1\% during training and 3\% for inference, which is insignificant on our hardware. ALiBi uses 100MB of extra memory when training on input lengths 1024 and 3072 in this setting. Memory usage is lower in all approaches when training on 3072 tokens (compared to 1024) since we break batches into multiple updates. See Table~\ref{tab:baselines_speed_mem} in the appendix for exact numbers. 
}
\label{fig:wt103_speed_mem}
\end{figure}

We show for the first time that the sinusoidal position method, which technically should be able to extrapolate, in practice has very limited extrapolation capabilities. Though the rotary position method improves over the sinusoidal one, it still does not achieve satisfying results.  Holding everything else constant, we are the first to observe that the T5 bias method leads to better extrapolation than either of these, and so we conclude that extrapolation ability depends heavily on the position embedding.  Unfortunately, the T5 bias is computationally costly (Figure~\ref{fig:wt103_speed_mem}).

\subsection{Background and Experimental Setup}

A transformer LM receives a list of tokens 
and outputs a probability distribution representing its prediction for the next token.  We call the input list the \textit{current input subsequence} since the inputs to language models are typically subsequences from (much longer) training or evaluation sequences.  During both training and perplexity evaluation (i.e., scoring a fixed sequence), many predictions can be calculated at once; this is done using a ``causal mask'' that ensures each position's prediction is influenced only by tokens to its left.  Let $L$ be the length of each input subsequence during training; it includes $L$ predictions, which on average have access to $\frac{L+1}{2}$ tokens of (left) context.  %
To explore a model's extrapolation abilities, we are interested in cases where sequences of length $L_{\textit{valid}} > L$ are considered at evaluation time. %
When $L$ differs between inference and training, we use \lt to refer to the length of subsequences during training and \li  to refer to their length at validation.

\paragraph{Nonoverlapping Inference}
To train on or evaluate a sequence longer than $L$ tokens, it is typical to segment the sequence into $L$-length subsequences and train on or evaluate them independently.  Unless otherwise stated, we use nonoverlapping inference to report perplexity scores. %

\paragraph{Extrapolation During Inference} 
Formally, the functions that define a transformer layer are agnostic to input length;\footnote{These include the embedding lookup, feedforward sublayer, and softmax layer, which act independently on vector inputs, as well as the attention sublayers, whose parameters do not depend on input length (and which must handle variable-length inputs, e.g., due to causal masking).  %
} they map from some arbitrary, unfixed number of input vectors to the same number of output vectors.  When transformers are applied to data that is inherently sequential, like text, %
positional information is injected into the inputs in various ways.

\citet{vaswani}
discussed two options for embedding positions into vectors to be added to word %
embeddings:  learning embeddings for specific positions and unlearned sinusoidal embeddings.  They observed similar performance between these two but preferred the sinusoidal approach, which they argued might extrapolate to longer input sequences during inference.  %
We find that this model cannot extrapolate to more than a few dozen tokens beyond $L$.\footnote{The learned positional embedding approach does not have a way to encode positions greater than $L$; it therefore has no ability to extrapolate.}

\paragraph{Experiment Setup}
We first test the extrapolation abilities of various position methods on the WikiText-103 corpus~\citep{pointer} using the transformer language model of~\cite{baevski}. 
We use this model because of its prominent role in recent  language modeling developments~\citep{khandelwal20generalization, shortformer}. The training set is about 103 million tokens from  English Wikipedia (half a gigabyte).  The  model has $16$ transformer layers of dimension $1024$, with $8$ heads, and a feedforward inner dimension of $4096$. This model ties the word embedding and softmax matrices~\citep{tying, inan2017}. %
In our experiments, other than varying the position method and training subsequence length, we modify no other hyperparameters, including the random seed and number of training epochs (205).

\subsection{Measuring Extrapolation}
\paragraph{Sinusoidal Position Embeddings}
Sinusoidal position embeddings (\citealp{vaswani}; \S 3.5) are constant, non-learned vectors that are added to token embeddings on input to the first layer of the transformer. They are frequently used in transformer language modeling~\citep{baevski,lewis2021base} and machine translation~\citep{vaswani,ott2018scaling} models. 
We first consider the unmodified model of~\cite{baevski}, which uses sinusoidal position embeddings, and train it on $L=512$ tokens; we then run inference with it on the validation set on $L+k$ tokens, with $k$ ranging from 0 to 15,000.
Figure~\ref{fig:wt103_extra} (left) and the corresponding Table~\ref{tab:appendix:wt103_extra_512} (in the appendix) show that while the model improves  perplexity up to $k=20$, performance stops improving and stays steady from $k=20$ to $k=50$ and then begins degrading. 
Similar results are obtained for a model trained with $L=1024$ tokens (Figure~\ref{fig:wt103_extra} (right) and Table~\ref{tab:appendix:wt103_extra_1024} in the appendix). That model improves for up to $\li = \lt + 50$ tokens, after which performance declines. 

\paragraph{Rotary Position Embeddings}
The rotary method was introduced by~\cite{roformer} and has recently been popularized by the open source GPT-3~\citep{gpt3} implementation GPT-J~\citep{gpt-j}. Instead of adding sinusoidal embeddings at the bottom of the transformer, they multiply the keys and queries of every attention layer by sinusoidal embeddings. 

Unlike the sinusoidal or learned positional embedding approach, the rotary method injects position information into the model at every layer, not just at the initial one.  In addition, it adds no position information to the values of the self-attention sublayer. The output of a self-attention sublayer is a linearly transformed, weighted sum of the input value vectors; therefore, by not inserting position information into the values, the outputs of each transformer-layer contain no explicit position information. 
We suspect that this segregation of position information may be beneficial for extrapolation, and we draw inspiration from it in the design of our method (\S\ref{sec:ourmethod}).

We apply the rotary position embedding method to our Baevski \& Auli baseline.\footnote{Our rotary method implementation is based on the code in \url{https://github.com/JunnYu/RoFormer_pytorch}, which is linked to from the official repository of~\cite{roformer}: (\url{https://github.com/ZhuiyiTechnology/roformer}). After we finished running our experiments with the rotary method, we were informed that the runtime of the code linked above could be optimized, making it only 2\% slower than the sinusoidal approach. This optimization would not change extrapolation performance.}
The perplexity results (Figure~\ref{fig:wt103_extra} and Appendix Tables~\ref{tab:appendix:wt103_extra_512} and~\ref{tab:appendix:wt103_extra_1024}) are better than the sinusoidal approach: the model with $L=512$ ($L=1024$) improves perplexity with up to $k=200$ ($k=100$) more tokens than it saw during training, but this comes at the cost of slower training and inference (Figure~\ref{fig:wt103_speed_mem}). %

\paragraph{T5 Bias}
Though most models use trained or sinusoidal position embeddings, the T5 model of~\cite{t5} uses a relative position method~\citep{shaw,Huang2019MusicTG} that adds no  position information to word embeddings (as in the previous method). Instead, it  %
modifies the way attention values are computed. We refer to this as the ``T5 bias'' method.\footnote{This method is similar to the one used in~\citet[Equation 7]{parikh-etal-2016-decomposable}.} To compute attention values in the unmodified transformer, we compute the dot product of every query with every relevant key %
and then softmax these attention values. In this method, we compute the attention values as before, %
but then we add a learned, shared bias to each query-key score that is dependent on just the distance between the query and key. Therefore, all query-key scores where the query and key distance are zero (i.e., the query and key represent the same token) get a specific learned bias, all scores where the query and key are one word away get a different learned bias, and so on, up to a certain point, from where multiple different distances share the same learned bias (which might be beneficial for extrapolation). %
As in the rotary method, the T5 bias injects position information into the model at every layer and integrates no explicit position information into the %
self-attention value vectors. 

\cite{t5} propose that the T5 bias may allow extrapolation, but they did not report experiments testing this.  Here, we show that the T5 bias does allow language models to extrapolate.
We do this by again modifying the Baevski \& Auli model, this time to insert the T5 bias into it.\footnote{Our T5 bias implementation is based on the one used in HuggingFace Transformers~\citep{huggingface}, which in turn is based on the official Mesh Tensorflow T5 code. }

As Figure~\ref{fig:wt103_extra} shows, the T5 bias improves perplexity with longer sequences than the ones it was trained on, i.e.,  $k=600$ ($k=800$) extra tokens for a model trained on $L=512$ ($L=1024$) input tokens.  Unfortunately, this impressive performance comes at a cost: training is at least twice as slow as with the sinusoidal model. Therefore, this model's extrapolation ability provides no efficiency advantage. For example, to do inference on 1024 tokens, we could either train the sinusoidal model with \lt = 1024 or train the T5 bias model on \lt = 512 tokens and extrapolate to 1024 for inference. However, the \lt = 1024 sinusoidal model runs at 28.5k words per second (WPS), while the \lt = 512 T5 bias model runs at 14.4k WPS (Appendix Table~\ref{tab:baselines_speed_mem}), so there is no speedup when training on shorter sequences with this method.\footnote{ \citet{narang2021transformer} benchmarked the T5 bias as being just 8.7\% slower than the sinusoidal approach; thus,   while always incurring a runtime penalty, this method's runtime could be faster depending on the choice of hardware and software frameworks used. Narang et al. used the Tensorflow T5 library running on TPUs, while we used the PyTorch Fairseq library running on GPUs. }

\section{Attention with Linear Biases (ALiBi)}
\label{sec:ourmethod}

\begin{figure}
\begin{center}
\includegraphics[width=.48\textwidth]{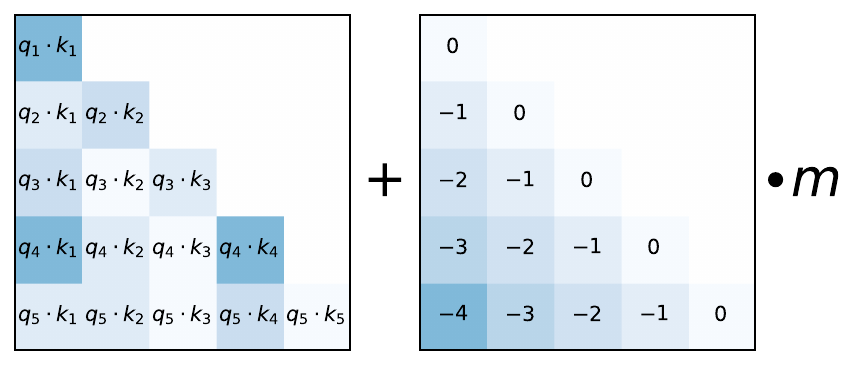} %
\end{center}
\caption{When computing attention scores for each head, our linearly biased attention method, ALiBi,  adds a constant bias (right) to each attention score ($\mathbf{q}_i \cdot \mathbf{k}_j$, left). As in the unmodified attention sublayer, the softmax function is then applied to these scores, and the rest of the computation is unmodified. \textbf{$\textbf{m}$ is a head-specific scalar} that is set and not learned throughout training. We show that our method for setting $m$ values generalizes to multiple text domains, models and training compute budgets. %
When using ALiBi, we do \emph{not} add positional embeddings at the bottom of the network.  }
\label{fig:1}
\end{figure}

In the transformer model of \citet{vaswani}, position embeddings are added to the word embeddings at the bottom of the network. %
For an input subsequence of length $L$, the attention sublayer %
computes the attention scores for the $i$th
query $\mathbf{q}_i \in \mathbb{R}^{1\times d}$, ($1 \leq i \leq L$) in each head, given the first $i$ keys $\mathbf{K} \in \mathbb{R}^{i\times d}$, where $d$ is the head dimension:%
\begin{equation*}
\text{softmax}(\mathbf{q}_i \mathbf{K}^\top)
\end{equation*}
These attention scores are then multiplied by the values to return the output of the attention sublayer.\footnote{For simplicity we omit the key, query, value and final output projections, dropout, and the scaling factor.}

When using ALiBi, we do not add position embeddings at any point in the network. %
The only modification we apply is after the query-key dot product, where we add a static, non-learned bias:\footnote{The ALiBi bias is not multiplied by the $\sqrt{d_k}$ scaling factor from Equation 1 of~\citet{vaswani}.}
\begin{equation*}
\text{softmax}(\mathbf{q}_i \mathbf{K}^\top + m \cdot [-(i-1), ..., -2, -1, 0]), 
\end{equation*}
where scalar $m$ is a head-specific slope fixed before training.
Figure~\ref{fig:1} offers a visualization. 

For our models with 8 heads, the slopes that we used are the geometric sequence: %
${\frac{1}{2^1}, \frac{1}{2^2}, ..., \frac{1}{2^8}}$. 
For models that require 16 heads, we interpolate those 8 slopes by geometrically averaging every consecutive pair, resulting in the geometric sequence that starts at $\frac{1}{\sqrt{2}}$ and has the ratio of $\frac{1}{\sqrt{2}}$: ${\frac{1}{2^{0.5}}, \frac{1}{2^1}, \frac{1}{2^{1.5}}, ..., \frac{1}{2^{8}}}$. In general, for $n$ heads, our set of slopes is the geometric sequence that starts at %
$2^{\frac{-8}{n}}$
and uses that same value as its ratio.

In \S\ref{sec:results}, we observe that this set of slopes works on a wide variety of text domains and model sizes. Therefore, we do not believe that it is necessary to tune these slope values every time a new model is trained on a new dataset. This makes our method similar to the sinusoidal approach, where the hyperparameters (the start and end of the geometric progression of wavelengths) were set once by~\citet{vaswani} and then reused in different models of different sizes on different datasets.

\al has an inductive bias towards recency; it penalizes attention scores between distant query-key pairs, with the penalty increasing as the distance between a key and a query grows. The different heads increase their penalties at different rates, depending on the slope magnitude. 

We initially experimented with making the slopes trainable, but this did not yield strong extrapolation results.\footnote{In our experiments, trainable slopes also slowed down the training speed by 3\%.} %
A brief manual exploration of around ten slope sets led us to discover the set of slopes that we finally picked. Our main insight from this exploration is that the slope sets that work best are those with slopes in the $(0,1)$ range, with the slopes' density increasing as we get closer to $0$. 
We also found our method to be robust to slope choice. Even randomly sampling from the exponential distribution worked well in some cases (although that method had high variance). %

Since ALiBi is a relative position method, we add position information at every layer to the keys and queries but not to the values, as is done in the T5 bias and rotary methods. We hypothesize that these properties might be beneficial for extrapolation. %

\paragraph{Implementation.}
ALiBi is easy to implement, with all changes accomplished in a few lines of code. We implement it by modifying the mask matrix by adding the linear biases to it (in practice, when training a transformer LM, query $\mathbf{q}_i$ attends only to keys $1$ to $i$; this is implemented by adding a mask matrix to the query-key dot product before the softmax operation is applied). This means that there is no runtime penalty when using our method since we add no operations to the network.%

Compared to the sinusoidal model trained on the same input lengths, AliBi incurs a memory increase (up to 100MB in some of our experiments): in the unmodified transformer, the mask is of size $L\times L$; when using ALiBi, the mask is a slightly larger $n \times L\times L$ (where $n$ is the number of heads) since the linear biases added for each head uses a different slope. But, as we show, ALiBi enables training on much smaller sequences while still achieving (and occasionally surpassing) results obtained using sinusoidal embeddings on longer sequences, which saves multiple gigabytes of memory. 

\section{Results}
\label{sec:results}
We first show that on WikiText103 ALiBi is efficient and enables training models with short input subsequences that outperform strong baselines even when the ALiBi models extrapolate to more than six times the number of tokens that they were trained on. 
We then take the same hyperparameters for our method (the set of slopes) that worked on WikiText-103 and show that -- with no modification -- they provide strong results on a dataset in a very different domain: books. 
Finally, we show that a 1.3B parameter model trained with AliBi on a much larger (461 GB) dataset with much more compute provides a superior alternative to the sinusoidal method since it achieves similar perplexity scores while running faster and using less memory (since it is trained on shorter inputs). 

While multiple alternatives to the position methods presented in~\cite{vaswani} have been proposed, few have been adopted in large (1B or more parameter) LMs since  that setting is much more challenging than the smaller scale experiments. GPT-3 and Jurassic-1~\citep{J1WhitePaper} use the learned position embedding method from Vaswani et al., and GPT-J uses the rotary method.
Our results on the 1.3B parameter model show our method's ability to generalize to larger models, dataset sizes and training durations without retuning the hyperparameter.

\subsection{Results on WikiText-103 and Toronto BookCorpus}

\begin{figure}[h]
\begin{center}
\includegraphics[width=.65\textwidth]{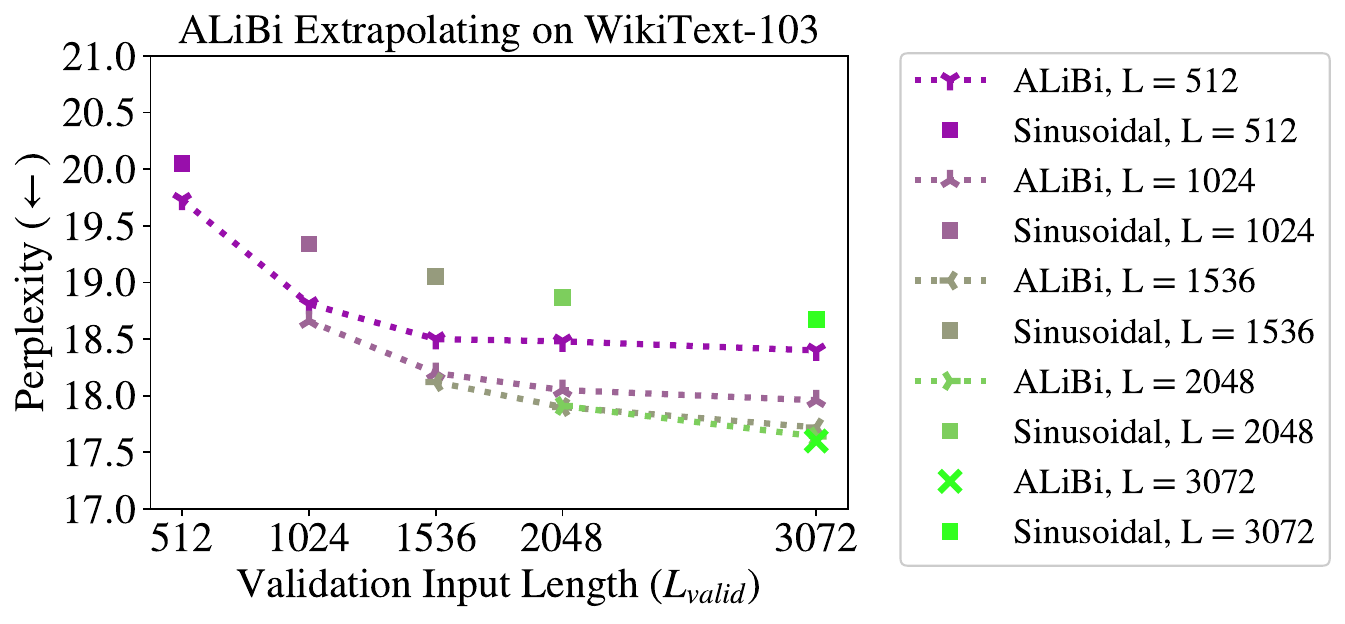} %
\end{center}
\caption{\al models trained and evaluated on varying sequence lengths on the WikiText-103 validation set and the sinusoidal baseline (not evaluated on longer sequences). All of our models outperform the sinusoidal ones even when trained on fewer tokens. Appendix Table~\ref{tab:LSB_wt103} has exact perplexities, more \al models (trained on fewer tokens), and results for rotary and T5 bias models. }
\label{fig:wt103_extra_lsb_all}
\end{figure}

We first develop our method on the WikiText-103 corpus~\citep{pointer}, replacing the sinusoidal position embeddings in the language model of~\cite{baevski} with \al.

Figure~\ref{fig:wt103_extra_lsb_all} (and the corresponding Appendix Table~\ref{tab:LSB_wt103}) show our results for models trained with varying numbers of input subsequence tokens ($L$), extrapolating to longer subsequence lengths on the validation dataset. 
Our first observation is that, without extrapolation, for every $L$, our models outperform  those using the sinusoidal method, sometimes by a  significant %
amount. For example, the Baevski \& Auli model achieves 18.67$\pm$0.24 (std.~dev.) perplexity when trained with $L = 3072$ input tokens, but our $L=3072$ model achieves 17.60 perplexity (when both models evaluate with \li = 3072).

Our second observation is that all of our models can extrapolate, and they obtain improved perplexity scores when handling more tokens than they observed during training. For example, our model trained on 512 tokens (which achieves 19.73 perplexity when evaluating subsequences of length 512 in the development set) achieves a perplexity score of 18.40 on the development set when extrapolating to subsequences of length 3072. Surprisingly, this surpasses the score that the $L=3072$ sinusoidal model obtains on the development set by a statistically significant margin. 
Note that all our models trained on $L=512$ to $L=2048$ outperform the sinusoidal baseline trained on $L=3072$ when extrapolating to \li = 3072 even though those models all take much less time to train since they train on shorter subsequences (Appendix Figure~\ref{fig:wt103_train_speed_vs_ppl} compares training speed to perplexity for these models)! The $L=512$ model is 1.84 times faster to train and yet still outperforms the $L=3072$ sinusoidal model when extrapolating to $\li = 3072$. In addition, training the $L=3072$ sinusoidal model requires a GPU with more than 16 GB of memory to fit the large attention matrices, which our $L=512$ outperforms even though it can be trained on a GPU with much less memory due to much smaller attention matrices. 

Additionally,  Table~\ref{tab:LSB_wt103} (in the appendix) also shows that, for $L$s  of 1024 and 3072, our method performs better than the rotary and T5 bias models even when \li = $L$ (i.e., no extrapolation is occurring).
Figure~\ref{fig:wt103_extra} (and the corresponding Appendix Tables~\ref{tab:appendix:wt103_extra_512} and~\ref{tab:appendix:wt103_extra_1024}) more broadly explore our method vs.~the other position methods. They show that the T5 bias (the best of the baselines) improves perplexity until \li is around $2L$, but on the WikiText-103 dataset our method continually improves perplexity until at least around $3L$, with the $L=512$ model improving perplexity even when \li exceeds 12k tokens. Even when unable to improve perplexity given longer sequences, \al always maintains strong performance as more tokens are added.

Appendix Table~\ref{tab:wt103_test} shows that our results on the validation set also transfer to the test set of WikiText-103.
Currently, almost all models that present results on WikiText-103 use sliding window evaluation  (defined in \S\ref{sec:analysis}) to compute perplexities. We apply that method to our (and to the sinusoidal, rotary and T5 bias) models in Appendix Table~\ref{tab:wt103_test_sota}. We find that our L = 3072 model surpasses the performance of Transformer-XL~\citep{transformer-xl}, the Sandwich~\citep{sandwich}, and Shortformer~\citep{shortformer} models. Our results are similar to the ones obtained with staged training~\citep{shortformer} but fall short of results obtained by Routing Transformer~\citep{roy2020efficient} and kNN-LM~\citep{khandelwal20generalization}. The methods used in those models are orthogonal to ours, and we hypothesize that combining them with ours might lead to even larger performance increases.

After developing our method on WikiText-103, in Appendix Section~\ref{subsec:tbc}, we run one set of experiments on a  different domain (books) using a similar model architecture and without modifying any of the \al hyperparameters (the slopes) and show that our results fully transfer to this new domain. Our models are able to both surpass the sinusoidal baseline when not extrapolating while also outperforming it when extrapolating to longer sequences.

\subsection{Results on the CC100+RoBERTa Corpus}

Our final set of experiments investigates whether \al transfers to a larger model trained with a larger computational budget on a larger dataset than the ones we previously used. We show that our method achieves strong results in this more challenging setting, obtaining similar performance to the sinusoidal baseline while using significantly less memory, since we train on shorter subsequences. 

The dataset we choose is a combination of the datasets used to train the RoBERTa~\citep{roberta} implementation of BERT~\citep{bert} and the English part of the CC-100 corpus introduced in~\cite{cc-100}, for a total of 461 GB. The RoBERTa training corpus---i.e., the Toronto Book Corpus~\citep{zhu2015aligning}, English Wikipedia, CC-News~\citep{ccnews}, OpenWebText~\citep{openwebtext} and Stories~\citep{stories})---is 161 gigabytes, and the English part of the CC-100 corpus is 300 gigabytes. 
The validation set contains 649K tokens.

Our models for this dataset have 25 transformer layers with 16 heads and a dimension of 2048, with an 8192 hidden dimension of the feedforward sublayers.
These models have 1.3B parameters. 
We train our models for one epoch, which is 50k updates on 128 V100 GPUs.

\begin{figure}
\centering
\begin{subfigure}{.45\textwidth} %
  \centering
  \includegraphics[width=\linewidth]{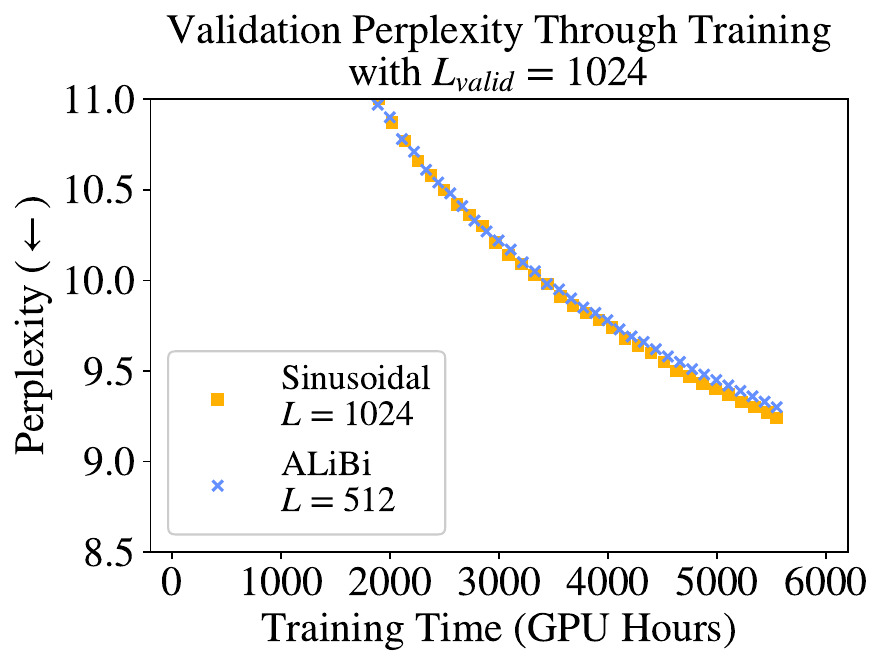}
 
\end{subfigure}%
\begin{subfigure}{.45\textwidth} %
  \centering
 \includegraphics[width=\linewidth]{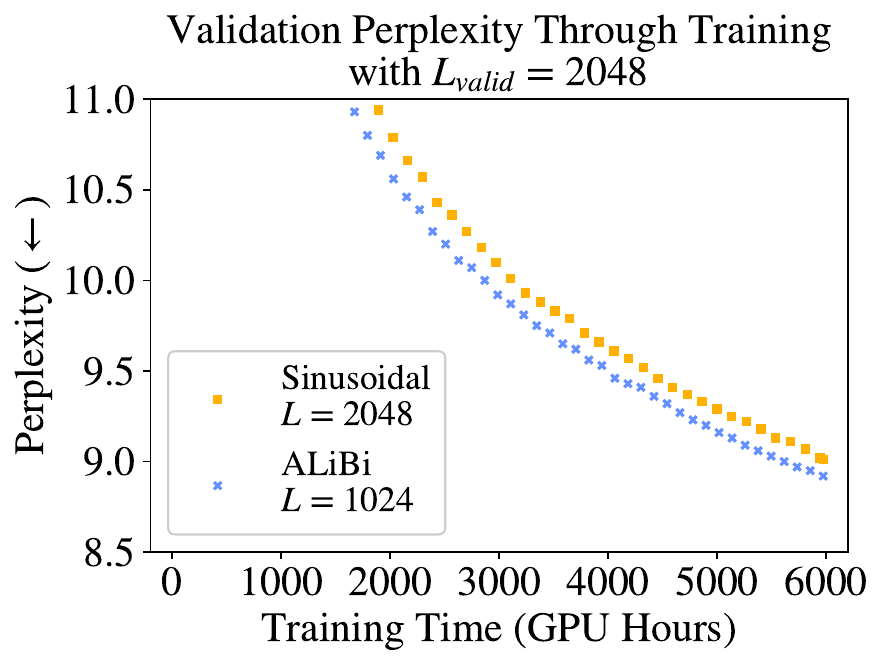}
\end{subfigure}
\caption{ On the left (right), a 1.3B-parameter ALiBi model trained on 512 (1024) and evaluated on 1024 (2048) tokens during training, compared to the sinusoidal baseline trained on 1024 (2048) tokens. The \al models obtain strong results even though they use 6\%-11\% less memory since they train on shorter sequences. Appendix Table \ref{tab:ccr} shows memory use and end-of-training perplexities.}
\label{fig:ccr_train_all}
\end{figure}

In Figure~\ref{fig:ccr_train_all} (left), we compare the validation perplexity for \li = 1024 throughout the training process for an \al model trained with \lt = 512 compared to the sinusoidal model trained with \lt = 1024. Since our model is trained on shorter sequences, it is 7\% faster and uses 1.6 GB less memory. We halt training of the sinusoidal baseline when our model reaches the end of its training (one epoch). 
At that time, our model is just 0.06 perplexity away from the baseline even though it was trained on sequences that are half the length of those the baseline used and requires less memory.

In Figure~\ref{fig:ccr_train_all} (right), results become even more impressive, showing that our model trained on \lt = 1024 outperforms by 0.09 perplexity the sinusoidal model trained on \lt = 2048 (when evaluating with \li = 2048) even though our model uses 3.1 GB less memory. Our model maintains a lead in perplexity over the sinusoidal model during the entire training process. By sampling five evenly distributed points across the training process, we compute that our \lt = 1024 model reaches a given perplexity value, on average, 11\% faster than the sinusoidal model does. 

Since our models in these comparisons use much less memory, they allow for stacking more layers, which would further improve performance (with negligible, if any, runtime cost). To keep our experiments as straightforward as possible, however, we do not add layers to our models.

Appendix Table~\ref{tab:ccr_all_50k} presents additional results comparing our models to the sinusoidal baseline when both are trained on the same $L$, showing that \al performs similarly to the sinusoidal baseline when not extrapolating. This contrasts with the  results presented on the smaller datasets, where \al consistently outperforms other position methods even when not extrapolating, suggesting that ALiBi's inductive bias provides additional benefits for lower-resource language modeling.

\begin{figure}[h]
\centering
\begin{subfigure}{.45\textwidth}  %
  \centering
  \includegraphics[width=\linewidth]{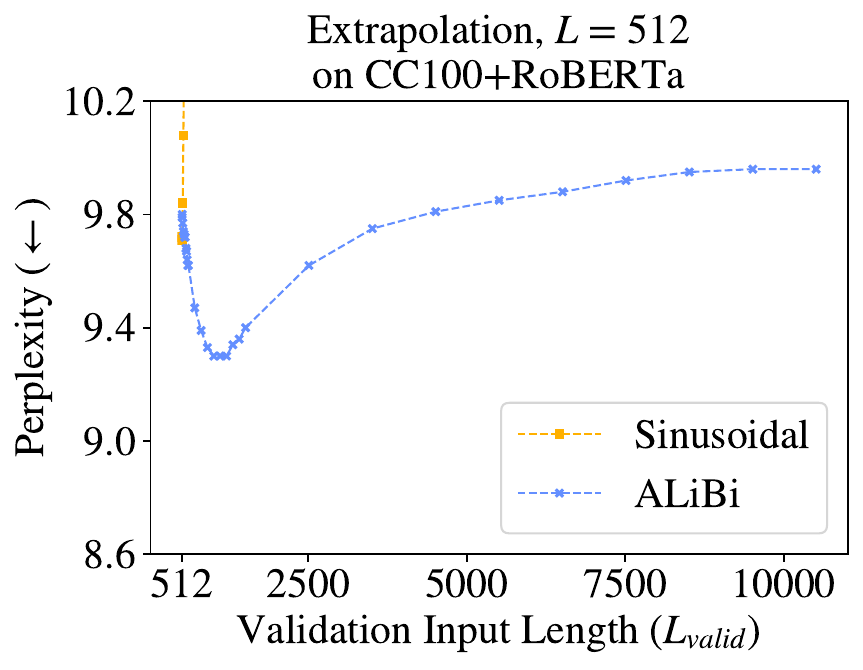}
\end{subfigure}%
\begin{subfigure}{.45\textwidth} %
  \centering
  \includegraphics[width=\linewidth]{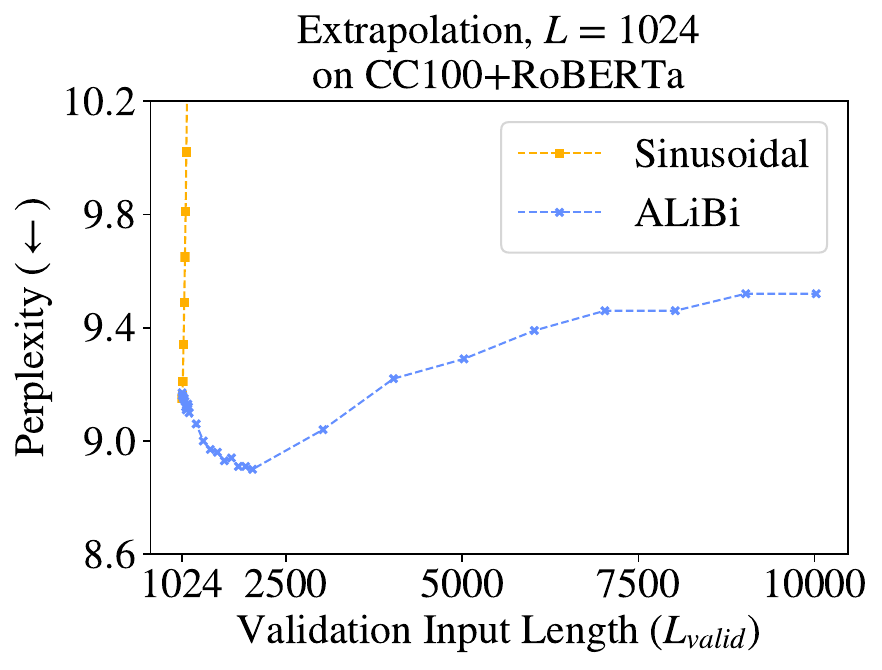}
\end{subfigure}
\caption{The ALiBi and sinusoidal models (with both $L$ = 512 and 1024) trained for 50k updates (1 epoch) on the CC100+RoBERTa corpus, extrapolating on the validation set. %
\al achieves the best results at around $2L$ but maintains strong performance even up to 10000 tokens in these experiments.}
\label{fig:ccr_extra}
\end{figure}

Figure~\ref{fig:ccr_extra} shows that our models trained on $L=512$ and $L=1024$ achieve the best results when extrapolating to about double the tokens that they were trained on.
Specifically, the \lt = 512 model (that obtains 9.79 perplexity when \li = 512) achieves its best score (9.3) when extrapolating to 1012 tokens, and the \lt = 1024 model (that obtains 9.16 perplexity when \li = 1024) achieves its best score (8.9) when extrapolating to 2024 tokens.

One possible explanation is that the subsequences the model observes during training are up to $L$ tokens long. When performing inference on subsequences of length $2L$, half of the subsequences the model consumes are as long as the examples seen during training. When inference is performed on subsequences of length $2L+1$ or longer, less than half of the predictions the model makes are on subsequences of lengths seen during training, and that might degrade performance. 

The sinusoidal model cannot extrapolate at all in this setting, with its performance degrading for both the \lt = 512 and 1024 models as soon as one token more than $L$ is added during evaluation.

In Appendix \ref{sec:analysis}, we find that \al's  edge over sinusoidal embeddings is largely explained by its improved avoidance of the early token curse.  We posit  that future work building on \al might achieve further gains by more efficiently  exploiting longer histories.

\section{Related Work}
In parallel with our work,~\citet{wennberg2021case} introduce a relative position method that, like our method, adds a bias to attention scores that is a function of the distance between the key and query elements. 
Unlike our \al method, which uses a  non-learned linear function, their method uses a radial-basis function, with multiple trainable parameters (in our experiments, this led to a slight decrease in runtime). 
In addition, they present experiments on text classification, not on language modeling.  They do not explore extrapolation.
The Distance Aware Transformer~\citep{da-transformer} multiplies attention scores by a bias that is a function of the distance between the key and query. This function uses a different, learned parameter in every head. They show results only on text classification. In our experiments (not presented), multiplying attention scores by the bias (instead of adding, as in \al) degraded performance. 

Transformer-XL~\citep{transformer-xl}  presented a language model that uses a cache and can attend to more tokens during inference than it was trained on (by increasing the length of the cache). However, this work presents results only where output length is limited to the $L$ (the training length), and their relative position method is very slow~\citep{shortformer}. 
The Longformer~\citep{longformer} adapts models trained on shorter sequences to document-level tasks. However, to achieve this they had to partially train their models on longer sequences. Our \al method enables extrapolation without any additional training on longer sequences. 

To our knowledge, extrapolation has not been previously explored in transformer language modeling, but it has been investigated previously and concurrently with transformers on other tasks, such as %
machine translation~\citep{rosendahl2019:pos_enc, neishi-yoshinaga-2019-relation, newman2020extrapolation, Kiyono2021SHAPESA}, sequence-to-sequence models trained on an artificial dataset~\citep{Hupkes2020}, pretrained sequence-to-sequence models tested on arithmetic tasks~\citep[Appendix C]{Nogueira2021InvestigatingTL}, models trained with reinforcement learning~\citep{lampinen2021towards}, image, speech recognition, and machine translation models~\citep{likhomanenko2021cape}, and protein structure prediction~\citep[Appendix 1.5]{Jumper2021HighlyAP}.

\section{Conclusion}

We showed that the sinusoidal position embedding approach does not enable transformers to extrapolate to inputs longer than the ones they were trained on. We then established that extrapolation in transformers can be enabled by just changing the position method.
We showed that our \al method offers an extremely simple replacement for existing position approaches and allow models to extrapolate. In addition, when not extrapolating, our method achieves either better perplexity than the sinusoidal method (in models smaller than 1B parameters, trained on less data) or similar perplexity (in larger, billion parameter models trained on much more data). 
\al is simple to implement and does not slow down runtime or require extra parameters (but does occasionally require a negligible amount of extra memory).
Using our method, we sped up the training of a 1.3 billion parameter model evaluated on the same input sequence length as GPT-3 (2048). 

\subsubsection*{Acknowledgments}
We thank Tim Dettmers, Gabriel Ilharco, Jungo Kasai, Hao Peng, Sewon Min, Sofia Serrano, Sam Shleifer, Luke Zettlemoyer, Julian Michael, Nikolaos Pappas, Yizhong Wang, and the anonymous reviewers for their valuable feedback and fruitful discussions.

\clearpage
\bibliography{iclr2022_conference}

\begin{thebibliography}{45}
\providecommand{\natexlab}[1]{#1}
\providecommand{\url}[1]{\texttt{#1}}
\expandafter\ifx\csname urlstyle\endcsname\relax
  \providecommand{\doi}[1]{doi: #1}\else
  \providecommand{\doi}{doi: \begingroup \urlstyle{rm}\Url}\fi

\bibitem[Baevski \& Auli(2018)Baevski and Auli]{baevski}
Alexei Baevski and Michael Auli.
\newblock Adaptive input representations for neural language modeling.
\newblock \emph{CoRR}, abs/1809.10853, 2018.
\newblock URL \url{http://arxiv.org/abs/1809.10853}.

\bibitem[Beltagy et~al.(2020)Beltagy, Peters, and Cohan]{longformer}
Iz~Beltagy, Matthew~E. Peters, and Arman Cohan.
\newblock Longformer: The long-document transformer.
\newblock \emph{arXiv:2004.05150}, 2020.

\bibitem[Brown et~al.(2020)Brown, Mann, Ryder, Subbiah, Kaplan, Dhariwal,
  Neelakantan, Shyam, Sastry, Askell, Agarwal, Herbert-Voss, Krueger, Henighan,
  Child, Ramesh, Ziegler, Wu, Winter, Hesse, Chen, Sigler, Litwin, Gray, Chess,
  Clark, Berner, McCandlish, Radford, Sutskever, and Amodei]{gpt3}
Tom~B. Brown, Benjamin Mann, Nick Ryder, Melanie Subbiah, Jared Kaplan,
  Prafulla Dhariwal, Arvind Neelakantan, Pranav Shyam, Girish Sastry, Amanda
  Askell, Sandhini Agarwal, Ariel Herbert-Voss, Gretchen Krueger, Tom Henighan,
  Rewon Child, Aditya Ramesh, Daniel~M. Ziegler, Jeffrey Wu, Clemens Winter,
  Christopher Hesse, Mark Chen, Eric Sigler, Mateusz Litwin, Scott Gray,
  Benjamin Chess, Jack Clark, Christopher Berner, Sam McCandlish, Alec Radford,
  Ilya Sutskever, and Dario Amodei.
\newblock Language models are few-shot learners, 2020.

\bibitem[Conneau et~al.(2020)Conneau, Khandelwal, Goyal, Chaudhary, Wenzek,
  Guzmán, Grave, Ott, Zettlemoyer, and Stoyanov]{cc-100}
Alexis Conneau, Kartikay Khandelwal, Naman Goyal, Vishrav Chaudhary, Guillaume
  Wenzek, Francisco Guzmán, Edouard Grave, Myle Ott, Luke Zettlemoyer, and
  Veselin Stoyanov.
\newblock Unsupervised cross-lingual representation learning at scale.
\newblock \emph{Proceedings of the 58th Annual Meeting of the Association for
  Computational Linguistics}, 2020.
\newblock \doi{10.18653/v1/2020.acl-main.747}.
\newblock URL \url{http://dx.doi.org/10.18653/v1/2020.acl-main.747}.

\bibitem[Dai et~al.(2019)Dai, Yang, Yang, Carbonell, Le, and
  Salakhutdinov]{transformer-xl}
Zihang Dai, Zhilin Yang, Yiming Yang, Jaime Carbonell, Quoc Le, and Ruslan
  Salakhutdinov.
\newblock Transformer-{XL}: Attentive language models beyond a fixed-length
  context.
\newblock In \emph{Proceedings of the 57th Annual Meeting of the Association
  for Computational Linguistics}, pp.\  2978--2988, Florence, Italy, July 2019.
  Association for Computational Linguistics.
\newblock \doi{10.18653/v1/P19-1285}.
\newblock URL \url{https://aclanthology.org/P19-1285}.

\bibitem[Devlin et~al.(2019)Devlin, Chang, Lee, and Toutanova]{bert}
Jacob Devlin, Ming-Wei Chang, Kenton Lee, and Kristina Toutanova.
\newblock {BERT}: Pre-training of deep bidirectional transformers for language
  understanding.
\newblock In \emph{Proceedings of the 2019 Conference of the North {A}merican
  Chapter of the Association for Computational Linguistics: Human Language
  Technologies, Volume 1 (Long and Short Papers)}, pp.\  4171--4186,
  Minneapolis, Minnesota, June 2019. Association for Computational Linguistics.
\newblock \doi{10.18653/v1/N19-1423}.
\newblock URL \url{https://www.aclweb.org/anthology/N19-1423}.

\bibitem[Gokaslan \& Cohen(2019)Gokaslan and Cohen]{openwebtext}
Aaron Gokaslan and Vanya Cohen.
\newblock Openwebtext corpus.
\newblock \url{http://Skylion007.github.io/OpenWebTextCorpus}, 2019.

\bibitem[Huang et~al.(2019)Huang, Vaswani, Uszkoreit, Simon, Hawthorne,
  Shazeer, Dai, Hoffman, Dinculescu, and Eck]{Huang2019MusicTG}
Cheng-Zhi~Anna Huang, Ashish Vaswani, Jakob Uszkoreit, Ian Simon, Curtis
  Hawthorne, Noam~M. Shazeer, Andrew~M. Dai, M.~Hoffman, M.~Dinculescu, and
  D.~Eck.
\newblock Music transformer: Generating music with long-term structure.
\newblock In \emph{ICLR}, 2019.

\bibitem[Hupkes et~al.(2020)Hupkes, Dankers, Mul, and Bruni]{Hupkes2020}
Dieuwke Hupkes, Verna Dankers, Mathijs Mul, and Elia Bruni.
\newblock Compositionality decomposed: How do neural networks generalise?
\newblock \emph{Journal of Artificial Intelligence Research}, 67:\penalty0
  757--795, April 2020.
\newblock \doi{10.1613/jair.1.11674}.
\newblock URL \url{https://doi.org/10.1613/jair.1.11674}.

\bibitem[Inan et~al.(2017)Inan, Khosravi, and Socher]{inan2017}
Hakan Inan, Khashayar Khosravi, and Richard Socher.
\newblock Tying word vectors and word classifiers: {A} loss framework for
  language modeling.
\newblock In \emph{ICLR}, 2017.
\newblock URL \url{https://openreview.net/forum?id=r1aPbsFle}.

\bibitem[Jumper et~al.(2021)Jumper, Evans, Pritzel, Green, Figurnov,
  Ronneberger, Tunyasuvunakool, Bates, Z{\'i}dek, Potapenko, Bridgland, Meyer,
  Kohl, Ballard, Cowie, Romera-Paredes, Nikolov, Jain, Adler, Back, Petersen,
  Reiman, Clancy, Zielinski, Steinegger, Pacholska, Berghammer, Bodenstein,
  Silver, Vinyals, Senior, Kavukcuoglu, Kohli, and
  Hassabis]{Jumper2021HighlyAP}
J.~Jumper, Richard Evans, A.~Pritzel, Tim Green, Michael Figurnov,
  O.~Ronneberger, Kathryn Tunyasuvunakool, Russ Bates, Augustin Z{\'i}dek, Anna
  Potapenko, A.~Bridgland, Clemens Meyer, Simon A~A Kohl, Andy Ballard,
  A.~Cowie, B.~Romera-Paredes, Stanislav Nikolov, Rishub Jain, J.~Adler,
  T.~Back, Stig Petersen, D.~Reiman, Ellen Clancy, Michal Zielinski, Martin
  Steinegger, Michalina Pacholska, Tamas Berghammer, S.~Bodenstein, D.~Silver,
  Oriol Vinyals, A.~Senior, K.~Kavukcuoglu, P.~Kohli, and D.~Hassabis.
\newblock Highly accurate protein structure prediction with alphafold.
\newblock \emph{Nature}, 596:\penalty0 583 -- 589, 2021.

\bibitem[Khandelwal et~al.(2020)Khandelwal, Levy, Jurafsky, Zettlemoyer, and
  Lewis]{khandelwal20generalization}
Urvashi Khandelwal, Omer Levy, Dan Jurafsky, Luke Zettlemoyer, and Mike Lewis.
\newblock {Generalization through Memorization: Nearest Neighbor Language
  Models}.
\newblock In \emph{International Conference on Learning Representations
  (ICLR)}, 2020.

\bibitem[Kiyono et~al.(2021)Kiyono, Kobayashi, Suzuki, and
  Inui]{Kiyono2021SHAPESA}
Shun Kiyono, Sosuke Kobayashi, Jun Suzuki, and Kentaro Inui.
\newblock Shape: Shifted absolute position embedding for transformers.
\newblock \emph{ArXiv}, abs/2109.05644, 2021.

\bibitem[Lampinen et~al.(2021)Lampinen, Chan, Banino, and
  Hill]{lampinen2021towards}
Andrew~Kyle Lampinen, Stephanie C.~Y. Chan, Andrea Banino, and Felix Hill.
\newblock Towards mental time travel: a hierarchical memory for reinforcement
  learning agents.
\newblock \emph{CoRR}, abs/2105.14039, 2021.
\newblock URL \url{https://arxiv.org/abs/2105.14039}.

\bibitem[Lewis et~al.(2021)Lewis, Bhosale, Dettmers, Goyal, and
  Zettlemoyer]{lewis2021base}
Mike Lewis, Shruti Bhosale, Tim Dettmers, Naman Goyal, and Luke Zettlemoyer.
\newblock Base layers: Simplifying training of large, sparse models, 2021.

\bibitem[Lieber et~al.(2021)Lieber, Sharir, Lenz, and Shoham]{J1WhitePaper}
Opher Lieber, Or~Sharir, Barak Lenz, and Yoav Shoham.
\newblock Jurassic-1: Technical details and evaluation.
\newblock Technical report, AI21 Labs, August 2021.

\bibitem[Likhomanenko et~al.(2021)Likhomanenko, Xu, Collobert, Synnaeve, and
  Rogozhnikov]{likhomanenko2021cape}
Tatiana Likhomanenko, Qiantong Xu, Ronan Collobert, Gabriel Synnaeve, and Alex
  Rogozhnikov.
\newblock {CAPE:} encoding relative positions with continuous augmented
  positional embeddings.
\newblock \emph{CoRR}, abs/2106.03143, 2021.
\newblock URL \url{https://arxiv.org/abs/2106.03143}.

\bibitem[Liu et~al.(2019)Liu, Ott, Goyal, Du, Joshi, Chen, Levy, Lewis,
  Zettlemoyer, and Stoyanov]{roberta}
Yinhan Liu, Myle Ott, Naman Goyal, Jingfei Du, Mandar Joshi, Danqi Chen, Omer
  Levy, Mike Lewis, Luke Zettlemoyer, and Veselin Stoyanov.
\newblock Roberta: A robustly optimized bert pretraining approach, 2019.

\bibitem[Merity et~al.(2016)Merity, Xiong, Bradbury, and Socher]{pointer}
Stephen Merity, Caiming Xiong, James Bradbury, and Richard Socher.
\newblock Pointer sentinel mixture models, 2016.

\bibitem[Mikolov \& Zweig(2012)Mikolov and Zweig]{Mikolov2012ContextDR}
Tomas Mikolov and G.~Zweig.
\newblock Context dependent recurrent neural network language model.
\newblock \emph{2012 IEEE Spoken Language Technology Workshop (SLT)}, pp.\
  234--239, 2012.

\bibitem[Mikolov et~al.(2010)Mikolov, Karafi{\'a}t, Burget, Cernock{\'y}, and
  Khudanpur]{Mikolov2010RecurrentNN}
Tomas Mikolov, M.~Karafi{\'a}t, L.~Burget, J.~Cernock{\'y}, and S.~Khudanpur.
\newblock Recurrent neural network based language model.
\newblock In \emph{INTERSPEECH}, 2010.

\bibitem[Nagel(2016)]{ccnews}
Sebastian Nagel.
\newblock Cc-news.
\newblock \url{https://commoncrawl.org/2016/10/news-dataset-available/}, 2016.

\bibitem[Narang et~al.(2021)Narang, Chung, Tay, Fedus, Fevry, Matena, Malkan,
  Fiedel, Shazeer, Lan, Zhou, Li, Ding, Marcus, Roberts, and
  Raffel]{narang2021transformer}
Sharan Narang, Hyung~Won Chung, Yi~Tay, William Fedus, Thibault Fevry, Michael
  Matena, Karishma Malkan, Noah Fiedel, Noam Shazeer, Zhenzhong Lan, Yanqi
  Zhou, Wei Li, Nan Ding, Jake Marcus, Adam Roberts, and Colin Raffel.
\newblock Do transformer modifications transfer across implementations and
  applications?, 2021.

\bibitem[Neishi \& Yoshinaga(2019)Neishi and
  Yoshinaga]{neishi-yoshinaga-2019-relation}
Masato Neishi and Naoki Yoshinaga.
\newblock On the relation between position information and sentence length in
  neural machine translation.
\newblock In \emph{Proceedings of the 23rd Conference on Computational Natural
  Language Learning (CoNLL)}, pp.\  328--338, Hong Kong, China, November 2019.
  Association for Computational Linguistics.
\newblock \doi{10.18653/v1/K19-1031}.
\newblock URL \url{https://aclanthology.org/K19-1031}.

\bibitem[Newman et~al.(2020)Newman, Hewitt, Liang, and
  Manning]{newman2020extrapolation}
Benjamin Newman, John Hewitt, Percy Liang, and Christopher~D. Manning.
\newblock The eos decision and length extrapolation.
\newblock In \emph{BlackBoxNLP@EMNLP}, 2020.
\newblock URL \url{https://nlp.stanford.edu/pubs/newman2020extrapolation.pdf}.

\bibitem[Nogueira et~al.(2021)Nogueira, Jiang, and
  Li]{Nogueira2021InvestigatingTL}
Rodrigo Nogueira, Zhiying Jiang, and Jimmy~J. Li.
\newblock Investigating the limitations of the transformers with simple
  arithmetic tasks.
\newblock \emph{ArXiv}, abs/2102.13019, 2021.

\bibitem[Ott et~al.(2018)Ott, Edunov, Grangier, and Auli]{ott2018scaling}
Myle Ott, Sergey Edunov, David Grangier, and Michael Auli.
\newblock Scaling neural machine translation.
\newblock In \emph{Proceedings of the Third Conference on Machine Translation
  (WMT)}, 2018.

\bibitem[Parikh et~al.(2016)Parikh, T{\"a}ckstr{\"o}m, Das, and
  Uszkoreit]{parikh-etal-2016-decomposable}
Ankur Parikh, Oscar T{\"a}ckstr{\"o}m, Dipanjan Das, and Jakob Uszkoreit.
\newblock A decomposable attention model for natural language inference.
\newblock In \emph{Proceedings of the 2016 Conference on Empirical Methods in
  Natural Language Processing}, pp.\  2249--2255, Austin, Texas, November 2016.
  Association for Computational Linguistics.
\newblock \doi{10.18653/v1/D16-1244}.
\newblock URL \url{https://aclanthology.org/D16-1244}.

\bibitem[Press \& Wolf(2017)Press and Wolf]{tying}
Ofir Press and Lior Wolf.
\newblock Using the output embedding to improve language models.
\newblock In \emph{Proceedings of the 15th Conference of the {E}uropean Chapter
  of the Association for Computational Linguistics: Volume 2, Short Papers},
  pp.\  157--163, Valencia, Spain, April 2017. Association for Computational
  Linguistics.
\newblock URL \url{https://www.aclweb.org/anthology/E17-2025}.

\bibitem[Press et~al.(2020)Press, Smith, and Levy]{sandwich}
Ofir Press, Noah~A. Smith, and Omer Levy.
\newblock Improving transformer models by reordering their sublayers.
\newblock In \emph{Proceedings of the 58th Annual Meeting of the Association
  for Computational Linguistics}, pp.\  2996--3005, Online, July 2020.
  Association for Computational Linguistics.
\newblock \doi{10.18653/v1/2020.acl-main.270}.
\newblock URL \url{https://www.aclweb.org/anthology/2020.acl-main.270}.

\bibitem[Press et~al.(2021)Press, Smith, and Lewis]{shortformer}
Ofir Press, Noah~A. Smith, and Mike Lewis.
\newblock Shortformer: Better language modeling using shorter inputs.
\newblock In \emph{Proceedings of the 59th Annual Meeting of the Association
  for Computational Linguistics and the 11th International Joint Conference on
  Natural Language Processing (Volume 1: Long Papers)}, pp.\  5493--5505,
  Online, August 2021. Association for Computational Linguistics.
\newblock URL \url{https://aclanthology.org/2021.acl-long.427}.

\bibitem[Rae et~al.(2020)Rae, Potapenko, Jayakumar, Hillier, and
  Lillicrap]{Rae2020Compressive}
Jack~W. Rae, Anna Potapenko, Siddhant~M. Jayakumar, Chloe Hillier, and
  Timothy~P. Lillicrap.
\newblock Compressive transformers for long-range sequence modelling.
\newblock In \emph{International Conference on Learning Representations}, 2020.
\newblock URL \url{https://openreview.net/forum?id=SylKikSYDH}.

\bibitem[Raffel et~al.(2020)Raffel, Shazeer, Roberts, Lee, Narang, Matena,
  Zhou, Li, and Liu]{t5}
Colin Raffel, Noam Shazeer, Adam Roberts, Katherine Lee, Sharan Narang, Michael
  Matena, Yanqi Zhou, Wei Li, and Peter~J. Liu.
\newblock Exploring the limits of transfer learning with a unified text-to-text
  transformer.
\newblock \emph{Journal of Machine Learning Research}, 21\penalty0
  (140):\penalty0 1--67, 2020.
\newblock URL \url{http://jmlr.org/papers/v21/20-074.html}.

\bibitem[Rosendahl et~al.(2019)Rosendahl, Tran, Wang, and
  Ney]{rosendahl2019:pos_enc}
Jan Rosendahl, Viet Anh~Khoa Tran, Weiyue Wang, and Hermann Ney.
\newblock Analysis of positional encodings for neural machine translation.
\newblock In \emph{International Workshop on Spoken Language Translation}, Hong
  Kong, China, November 2019.

\bibitem[Roy et~al.(2020)Roy, Saffar, Vaswani, and Grangier]{roy2020efficient}
Aurko Roy, Mohammad Saffar, Ashish Vaswani, and David Grangier.
\newblock Efficient content-based sparse attention with routing transformers,
  2020.

\bibitem[Shaw et~al.(2018)Shaw, Uszkoreit, and Vaswani]{shaw}
Peter Shaw, Jakob Uszkoreit, and Ashish Vaswani.
\newblock Self-attention with relative position representations.
\newblock In \emph{Proceedings of the 2018 Conference of the North {A}merican
  Chapter of the Association for Computational Linguistics: Human Language
  Technologies, Volume 2 (Short Papers)}, pp.\  464--468, New Orleans,
  Louisiana, June 2018. Association for Computational Linguistics.
\newblock \doi{10.18653/v1/N18-2074}.
\newblock URL \url{https://www.aclweb.org/anthology/N18-2074}.

\bibitem[Su et~al.(2021)Su, Lu, Pan, Wen, and Liu]{roformer}
Jianlin Su, Yu~Lu, Shengfeng Pan, Bo~Wen, and Yunfeng Liu.
\newblock Roformer: Enhanced transformer with rotary position embedding, 2021.

\bibitem[Trinh \& Le(2018)Trinh and Le]{stories}
Trieu~H. Trinh and Quoc~V. Le.
\newblock A simple method for commonsense reasoning, 2018.

\bibitem[Vaswani et~al.(2017)Vaswani, Shazeer, Parmar, Uszkoreit, Jones, Gomez,
  Kaiser, and Polosukhin]{vaswani}
Ashish Vaswani, Noam Shazeer, Niki Parmar, Jakob Uszkoreit, Llion Jones,
  Aidan~N Gomez, \L~ukasz Kaiser, and Illia Polosukhin.
\newblock Attention is all you need.
\newblock In I.~Guyon, U.~V. Luxburg, S.~Bengio, H.~Wallach, R.~Fergus,
  S.~Vishwanathan, and R.~Garnett (eds.), \emph{Advances in Neural Information
  Processing Systems}, volume~30. Curran Associates, Inc., 2017.
\newblock URL
  \url{https://proceedings.neurips.cc/paper/2017/file/3f5ee243547dee91fbd053c1c4a845aa-Paper.pdf}.

\bibitem[Wang \& Komatsuzaki(2021)Wang and Komatsuzaki]{gpt-j}
Ben Wang and Aran Komatsuzaki.
\newblock {GPT-J-6B: A 6 Billion Parameter Autoregressive Language Model}.
\newblock \url{https://github.com/kingoflolz/mesh-transformer-jax}, May 2021.

\bibitem[Wennberg \& Henter(2021)Wennberg and Henter]{wennberg2021case}
Ulme Wennberg and Gustav~Eje Henter.
\newblock The case for translation-invariant self-attention in
  transformer-based language models, 2021.

\bibitem[Wolf et~al.(2020)Wolf, Debut, Sanh, Chaumond, Delangue, Moi, Cistac,
  Rault, Louf, Funtowicz, Davison, Shleifer, von Platen, Ma, Jernite, Plu, Xu,
  Scao, Gugger, Drame, Lhoest, and Rush]{huggingface}
Thomas Wolf, Lysandre Debut, Victor Sanh, Julien Chaumond, Clement Delangue,
  Anthony Moi, Pierric Cistac, Tim Rault, Rémi Louf, Morgan Funtowicz, Joe
  Davison, Sam Shleifer, Patrick von Platen, Clara Ma, Yacine Jernite, Julien
  Plu, Canwen Xu, Teven~Le Scao, Sylvain Gugger, Mariama Drame, Quentin Lhoest,
  and Alexander~M. Rush.
\newblock Transformers: State-of-the-art natural language processing.
\newblock In \emph{Proceedings of the 2020 Conference on Empirical Methods in
  Natural Language Processing: System Demonstrations}, pp.\  38--45, Online,
  October 2020. Association for Computational Linguistics.
\newblock URL \url{https://www.aclweb.org/anthology/2020.emnlp-demos.6}.

\bibitem[Wu et~al.(2021)Wu, Wu, and Huang]{da-transformer}
Chuhan Wu, Fangzhao Wu, and Yongfeng Huang.
\newblock {DA}-transformer: Distance-aware transformer.
\newblock In \emph{Proceedings of the 2021 Conference of the North American
  Chapter of the Association for Computational Linguistics: Human Language
  Technologies}, pp.\  2059--2068, Online, June 2021. Association for
  Computational Linguistics.
\newblock \doi{10.18653/v1/2021.naacl-main.166}.
\newblock URL \url{https://aclanthology.org/2021.naacl-main.166}.

\bibitem[Zaremba et~al.(2014)Zaremba, Sutskever, and
  Vinyals]{zaremba2014recurrent}
Wojciech Zaremba, Ilya Sutskever, and Oriol Vinyals.
\newblock Recurrent neural network regularization, 2014.

\bibitem[Zhu et~al.(2015)Zhu, Kiros, Zemel, Salakhutdinov, Urtasun, Torralba,
  and Fidler]{zhu2015aligning}
Yukun Zhu, Ryan Kiros, Rich Zemel, Ruslan Salakhutdinov, Raquel Urtasun,
  Antonio Torralba, and Sanja Fidler.
\newblock Aligning books and movies: Towards story-like visual explanations by
  watching movies and reading books.
\newblock In \emph{Proceedings of the IEEE international conference on computer
  vision}, pp.\  19--27, 2015.

\end{thebibliography}
\bibliographystyle{iclr2022_conference}

\clearpage
\appendix
\section{Appendix}
\label{sec:appendix}

\subsection{Introduction}

\begin{figure}[h]
\begin{center}
\includegraphics[width=1\textwidth]{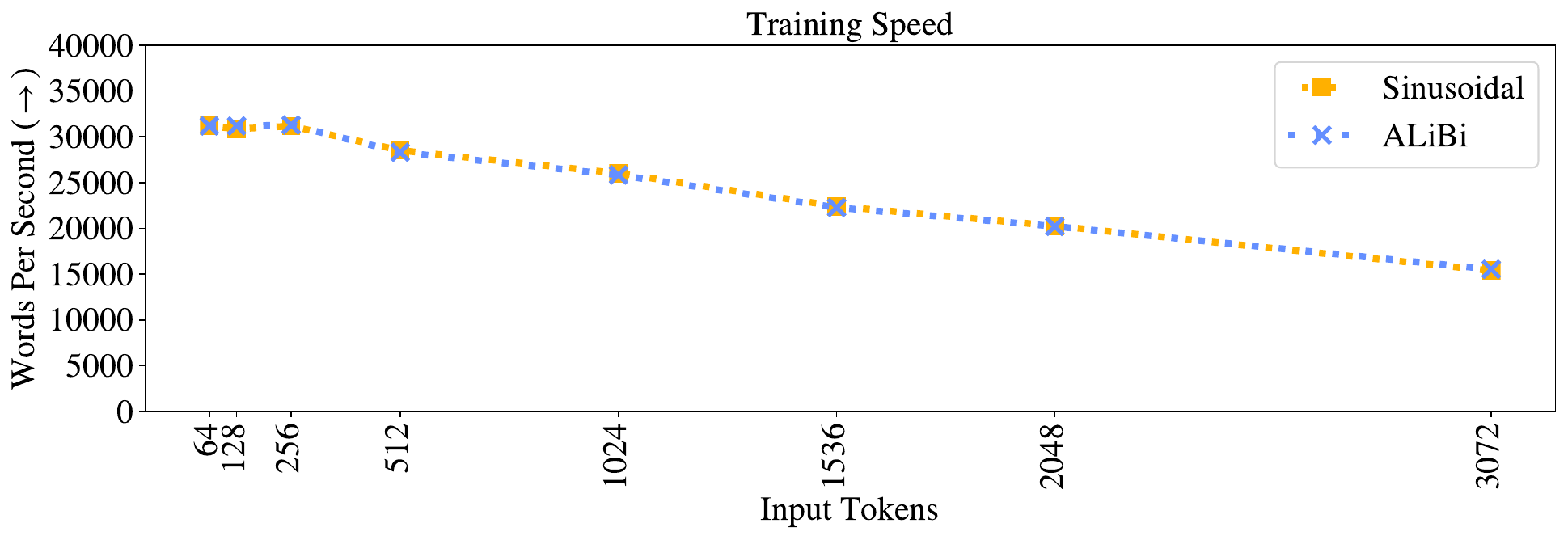}
\end{center}
\caption{ Training speed of our model and the sinusoidal baseline trained on different amounts of input subsequence tokens $L$.}
\label{fig:wt103_train_speeds_all}
\end{figure}

The training speed of transformer LMs gets slower as the input subsequence length $L$ increases. Figure~\ref{fig:wt103_train_speeds_all} visualizes this.

Table~\ref{tab:baselines_speed_mem} contains the runtimes and memory use statistics for models using the various position methods discussed in this work. 
\begin{table}[h]

\caption{ The speed (during training and evaluation, in words per second) and memory usage (during training) of the rotary, T5 bias, and ALiBi models compared to the sinusoidal baseline  on WikiText-103. Training and inference are batched, and speeds are shown for one V100 GPU. }
\label{tab:baselines_speed_mem}
\begin{center}
\begin{tabular}{@{}lrccc@{}}
\toprule
\multirow{2}{*}[0pt]{Position Method}& \multirow{2}{*}[0pt]{Train Length}&  \multicolumn{2}{c}{Speed ($\uparrow$)} & \multirow{2}{*}[0pt]{Memory ($\downarrow$)} \\
                &  & Train  & Eval. &   \\\midrule

& 512           &28.5k &82.1k  & 15.3 GB  \\
Sinusoidal&1024 &26.0k &77.8k  & 19.2 GB  \\
& 3072          &15.3k &42.4k  & 15.1 GB \\\midrule
&512            &20.0k &43.4k  & 17.8 GB  \\
Rotary & 1024   &17.7k &39.4k  & 22.8 GB  \\
& 3072          &11.5k &29.5k  & 17.8 GB  \\\midrule
& 512           &14.4k &21.8k  & 16.9 GB  \\
T5 Bias& 1024   &13.0k &20.2k  & 20.9 GB  \\
& 3072          & 4.3k & 4.9k  & 15.9 GB  \\\midrule
& 512           &28.3k &85.8k  & 15.3 GB \\
ALiBi & 1024    &25.8k &76.4k  & 19.3 GB \\
& 3072          &15.5k &42.2k  & 15.2 GB \\

\bottomrule
\end{tabular}
\end{center}
\end{table}

Tables~\ref{tab:appendix:wt103_extra_512}, \ref{tab:appendix:wt103_extra_1024}, and \ref{tab:appendix:wt103_extra_3072} show the perplexity and runtime of models using the sinusoidal, rotary T5 bias, and \al position methods when extrapolating to sequences longer than the ones they were trained on. The models used in these tables were trained on $L = 512, 1024$ and $3072$ tokens. 
\begin{table}[]
\small
\caption{The sinusoidal, rotary, T5 bias and \al models trained on \lt = \textbf{512} on WikiText-103 and evaluated with different values of \li on the validation set. \textbf{Bold} shows the best score for each model. Inference speeds (in words per second) are from inference on a GPU with batch size of one.}
\label{tab:appendix:wt103_extra_512}
\begin{center}

\begin{tabular}{@{}lcccccccc@{}} \toprule
 &  \multicolumn{2}{c}{Sinusoidal} & \multicolumn{2}{c}{Rotary} &\multicolumn{2}{c}{T5 Bias} & \multicolumn{2}{c}{ALiBi} \\
Inputs & PPL ($\downarrow$) & WPS ($\uparrow$)& PPL ($\downarrow$) & WPS ($\uparrow$)& PPL ($\downarrow$) & WPS ($\uparrow$)& PPL ($\downarrow$) & WPS ($\uparrow$) \\ \cmidrule(r){1-1} \cmidrule(l){2-9}

512    & 20.05                           & 15046 & 20.07                           & 10839 & 19.65                           & 11724 & 19.73                           & 14726 \\
513    & 19.98                           & 14925 & 20.01                           & 10806 & 19.57                           & 10491 & 19.62                           & 14965 \\
522    & 19.93                           & 15116 & 20.02                           & 11295 & 19.57                           & 9970  & 19.64                           & 15316 \\
532    & \textbf{19.91} & 15358 & 19.98                           & 10854 & 19.53                           & 10382 & 19.61                           & 15383 \\
542    & \textbf{19.91} & 15076 & 19.94                           & 10795 & 19.47                           & 12270 & 19.57                           & 15301 \\
552    & \textbf{19.91} & 16394 & 19.93                           & 12267 & 19.47                           & 13000 & 19.54                           & 16540 \\
562    & \textbf{19.91} & 16646 & 19.87                           & 12481 & 19.39                           & 12201 & 19.49                           & 16385 \\
572    & 19.95                           & 16934 & 19.83                           & 12668 & 19.36                           & 12851 & 19.46                           & 16881 \\
582    & 20.13                           & 16961 & 19.88                           & 12594 & 19.41                           & 13904 & 19.48                           & 17064 \\
592    & 20.18                           & 17243 & 19.84                           & 13007 & 19.36                           & 13706 & 19.43                           & 17289 \\
602    & 20.40                           & 17502 & 19.81                           & 12788 & 19.33                           & 14102 & 19.38                           & 17141 \\
612    & 20.59                           & 17637 & 19.81                           & 12601 & 19.27                           & 14573 & 19.38                           & 17661 \\
712    & 24.86                           & 15614 & \textbf{19.79} & 12676 & 19.10                           & 13818 & 19.14                           & 15637 \\
812    & 30.82                           & 17151 & 20.17                           & 13954 & 18.94                           & 14377 & 18.99                           & 17210 \\
912    & 37.42                           & 17200 & 20.73                           & 13887 & 18.86                           & 15345 & 18.88                           & 17619 \\
1012   & 43.54                           & 16304 & 21.37                           & 13759 & 18.79                           & 14240 & 18.73                           & 16059 \\
1112   & 50.36                           & 16424 & 22.01                           & 13891 & \textbf{18.77} & 14014 & 18.68                           & 16659 \\
1212   & 58.01                           & 17294 & 23.02                           & 15245 & 18.87                           & 14589 & 18.67                           & 17372 \\
1312   & 63.62                           & 15314 & 23.93                           & 13698 & 18.84                           & 13138 & 18.60                           & 15698 \\
1412   & 70.75                           & 15663 & 24.81                           & 13928 & 18.87                           & 12857 & 18.59                           & 15860 \\
1512   & 76.23                           & 15812 & 25.99                           & 14248 & 18.91                           & 13752 & 18.52                           & 16225 \\
2512   & 132.41                          & 15254 & 31.58                           & 13456 & 20.41                           & 9948  & 18.41                           & 15204 \\
3512   & 178.97                          & 13293 & 35.54                           & 11850 & 22.91                           & 7847  & 18.40                           & 13329 \\
4512   & 209.37                          & 11767 & 39.15                           & 10485 & 25.91                           & 6146  & 18.41                           & 11738 \\
5512   & 240.44                          & 10168 & 43.14                           & 9020  & 29.54                           & 5309  & 18.36                           & 9986  \\
6512   & 271.40                          & 9052  & 47.81                           & 8108  & 34.48                           & 4680  & 18.35                           & 9022  \\
7512   & 293.02                          & 8315  & 51.12                           & 7483  & 39.29                           & 4102  & 18.33                           & 8324  \\
8512   & 305.65                          & 7259  & 54.98                           & 6718  & 43.08                           & 3660  & 18.34                           & 7366  \\
9512   & 336.02                          & 6672  & 57.85                           & 6211  & 48.90                           & 3370  & 18.34                           & 6555  \\
10512  & 341.53                          & 6126  & 60.77                           & 5575  & 52.95                           & 3010  & 18.32                           & 6030  \\
11512  & 362.74                          & 5994  & 66.62                           & 5445  & 61.38                           & 2873  & 18.32                           & 5882  \\
12512  & 373.17                          & 5421  & 69.70                           & 4988  & 64.94                           & 2602  & \textbf{18.31} & 5287  \\
13512  & 382.91                          & 5174  & 73.27                           & 4692  & OOM                             & -     & \textbf{18.31} & 4962  \\
14512  & 399.98                          & 4351  & 75.52                           & 4103  & OOM                             & -     & \textbf{18.31} & 4352  \\
15512  & 406.01                          & 4291  & 79.25                           & 3969  & OOM                             & -     & \textbf{18.31} & 4289  \\

 \bottomrule
\end{tabular}
\end{center}
\end{table}

\begin{table}[]
\small
\caption{The sinusoidal, rotary, T5 bias and \al models  trained on \lt = \textbf{1024} on WikiText-103 and evaluated with different values of \li on the validation set. \textbf{Bold} shows the best score for each model. Inference speeds (in words per second) are from inference on a GPU with batch size of one.}
\label{tab:appendix:wt103_extra_1024}
\begin{center}

\begin{tabular}{@{}lcccccccc@{}} \toprule
 &  \multicolumn{2}{c}{Sinusoidal} & \multicolumn{2}{c}{Rotary} &\multicolumn{2}{c}{T5 Bias} & \multicolumn{2}{c}{ALiBi} \\
Inputs & PPL ($\downarrow$) & WPS ($\uparrow$)& PPL ($\downarrow$) & WPS ($\uparrow$)& PPL ($\downarrow$) & WPS ($\uparrow$)& PPL ($\downarrow$) & WPS ($\uparrow$) \\ \cmidrule(r){1-1} \cmidrule(l){2-9}
1024   & 19.34      & 17002 & 19.33  & 14690 & 18.80   & 14973 & 18.66 & 16951 \\
1025   & 19.33      & 16630 & 19.34  & 14423 & 18.82   & 14635 & 18.67 & 16690 \\
1034   & 19.27      & 16589 & 19.28  & 14351 & 18.74   & 14435 & 18.60 & 16707 \\
1044   & 19.26      & 16760 & 19.27  & 14491 & 18.72   & 14644 & 18.60 & 16667 \\
1054   & 19.23      & 16747 & 19.26  & 14503 & 18.71   & 14800 & 18.58 & 16833 \\
1064   & 19.21      & 16676 & 19.22  & 14623 & 18.70   & 14498 & 18.55 & 16941 \\
1074   &\textbf{ 19.19}      & 16879 & 19.19  & 14464 & 18.65   & 14670 & 18.49 & 16936 \\
1084   & 19.22      & 16942 & 19.23  & 14650 & 18.70   & 14607 & 18.56 & 17090 \\
1094   & 19.24      & 16771 & 19.22  & 14629 & 18.69   & 14517 & 18.54 & 16880 \\
1104   & 19.28      & 16870 & 19.27  & 14837 & 18.69   & 14635 & 18.52 & 17009 \\
1114   & 19.29      & 16795 & 19.27  & 14879 & 18.69   & 14540 & 18.52 & 17050 \\
1124   & 19.26      & 17312 & \textbf{19.18}  & 15121 & 18.62   & 14480 & 18.46 & 17571 \\
1224   & 20.54      & 17901 & 19.38  & 15584 & 18.58   & 14956 & 18.40 & 18013 \\
1324   & 23.13      & 16308 & 19.96  & 14386 & 18.52   & 13726 & 18.33 & 16422 \\
1424   & 26.45      & 16217 & 21.27  & 14385 & 18.48   & 13516 & 18.28 & 16121 \\
1524   & 29.82      & 16377 & 22.59  & 14693 & 18.42   & 13587 & 18.22 & 16659 \\
1624   & 34.27      & 15928 & 24.34  & 14228 & 18.40   & 12979 & 18.17 & 16053 \\
1724   & 38.24      & 16640 & 25.66  & 14686 & 18.35   & 12976 & 18.15 & 16607 \\
1824   & 42.23      & 16840 & 27.63  & 14918 & \textbf{18.30}   & 13071 & 18.08 & 16846 \\
1924   & 46.46      & 15071 & 29.64  & 13452 & 18.31   & 11843 & 18.08 & 15118 \\
2024   & 51.09      & 15591 & 31.17  & 13706 & 18.34   & 11906 & 18.05 & 15557 \\
3024   & 96.46      & 13639 & 35.67  & 12256 & 18.62   & 8480  & \textbf{17.92} & 13668 \\
4024   & 144.00     & 12441 & 44.30  & 11203 & 19.44   & 7443  & 17.95 & 12402 \\
5024   & 182.31     & 11431 & 48.31  & 10324 & 20.47   & 6384  &\textbf{ 17.92} & 11394 \\
6024   & 214.02     & 10238 & 54.78  & 9117  & 21.76   & 5577  & 18.01 & 10119 \\
7024   & 261.86     & 8785  & 62.83  & 7950  & 23.64   & 4867  & 17.93 & 8779  \\
8024   & 284.88     & 8132  & 64.91  & 7355  & 25.79   & 4377  & 17.96 & 8086  \\
9024   & 310.04     & 7045  & 71.91  & 6380  & 27.54   & 3787  & 17.98 & 7001  \\
10024  & 337.48     & 6633  & 77.70  & 6016  & 29.54   & 3582  & 17.97 & 6583  \\
11024  & 358.43     & 5722  & 81.15  & 5219  & 31.94   & 3170  & 18.02 & 5641  \\
12024  & 375.95     & 5560  & 87.51  & 5072  & 33.35   & 2940  & 18.01 & 5294  \\
13024  & 393.57     & 4691  & 94.74  & 4383  & OOM     & -      & 17.98 & 4621  \\
14024  & 403.52     & 4905  & 96.10  & 4546  & OOM     &  -     & 18.01 & 4827  \\
15024  & 431.66     & 4518  & 99.78  & 4170  & OOM     &   -    & 17.96 & 4447  \\
16024  & 453.32     & 4239  & 106.99 & 3878  & OOM     &   -    & 17.98 & 4153  \\ 
 \bottomrule
\end{tabular}
\end{center}
\end{table}

\begin{table}[]
\small
\caption{The sinusoidal, rotary, T5 bias and \al models  trained on \lt = \textbf{3072} on WikiText-103 and evaluated with different values of \li on the validation set. \textbf{Bold} shows the best score for each model. Inference speeds (in words per second) are from inference on a GPU with batch size of one.}
\label{tab:appendix:wt103_extra_3072}
\begin{center}

\begin{tabular}{@{}lcccccccc@{}} \toprule
 &  \multicolumn{2}{c}{Sinusoidal} & \multicolumn{2}{c}{Rotary} &\multicolumn{2}{c}{T5 Bias} & \multicolumn{2}{c}{ALiBi} \\
Inputs & PPL ($\downarrow$) & WPS ($\uparrow$)& PPL ($\downarrow$) & WPS ($\uparrow$)& PPL ($\downarrow$) & WPS ($\uparrow$)& PPL ($\downarrow$) & WPS ($\uparrow$) \\ \cmidrule(r){1-1} \cmidrule(l){2-9}
3072  & 18.67  & 13380 & 18.57 & 12548 & 18.01 & 8828 & 17.60 & 13866 \\ 
3073  & 18.67  & 13773 & 18.57 & 12474 & 18.01 & 8483 & 17.59 & 13793 \\
3082  & 18.62  & 13741 & 18.54 & 12388 & 17.95 & 8698 & 17.59 & 13778 \\
3092  &\textbf{ 18.60 } & 13742 & 18.48 & 12458 & 17.92 & 8361 & 17.55 & 13783 \\
3102  & 18.65  & 13701 & 18.52 & 12365 & 17.94 & 8764 & 17.59 & 13747 \\
3112  & 18.64  & 13809 & 18.51 & 12449 & 17.96 & 8665 & 17.59 & 13827 \\
3122  & 18.68  & 13722 & 18.52 & 12432 & 17.98 & 8437 & 17.58 & 13795 \\
3132  & 18.67  & 13825 & 18.54 & 12490 & 17.97 & 8653 & 17.58 & 13784 \\
3142  & 18.69  & 13543 & 18.52 & 12230 & 17.97 & 8282 & 17.61 & 13572 \\
3152  & 18.66  & 13520 & 18.56 & 12240 & 17.98 & 8608 & 17.59 & 13523 \\
3162  & 18.71  & 13501 & 18.56 & 12253 & 18.04 & 8589 & 17.62 & 13598 \\
3172  & 18.72  & 13563 & 18.55 & 12297 & 17.99 & 8583 & 17.59 & 13625 \\
3272  & 18.87  & 13453 & 18.55 & 12148 & 17.93 & 8144 & 17.59 & 13482 \\
3372  & 19.46  & 13533 & 18.50 & 12254 & 17.88 & 8442 & 17.52 & 13565 \\
3472  & 20.55  & 13047 & 18.52 & 11868 & 17.95 & 7857 & 17.54 & 13107 \\
3572  & 21.84  & 13128 & 18.50 & 11882 & 17.86 & 7814 & 17.50 & 13170 \\
3672  & 23.04  & 13106 & 18.49 & 11859 & 17.87 & 7719 & 17.48 & 13196 \\
3772  & 24.47  & 13287 & 18.54 & 11942 & 17.85 & 7579 & 17.49 & 13312 \\
3872  & 25.85  & 12621 & \textbf{18.40} & 11272 & 17.82 & 7581 & 17.41 & 12566 \\
3972  & 27.21  & 12379 & 18.48 & 11151 & 17.84 & 7483 & 17.41 & 12324 \\
4072  & 28.59  & 12178 & 18.59 & 11019 & 17.88 & 6974 & 17.48 & 12212 \\
5072  & 45.53  & 11076 & 18.80 & 9887  & 17.76 & 6230 & 17.33 & 10938 \\
6072  & 65.01  & 10114 & 19.50 & 9049  &\textbf{ 17.68} & 5554 & 17.26 & 10133 \\
7072  & 85.96  & 8647  & 20.60 & 7861  & 17.83 & 4820 & 17.22 & 8670  \\
8072  & 102.74 & 7755  & 21.60 & 6991  & 18.06 & 4281 & 17.30 & 7729  \\
9072  & 125.99 & 6953  & 22.14 & 6360  & 18.12 & 3823 & 17.26 & 6939  \\
10072 & 133.68 & 6646  & 23.21 & 6068  & 18.37 & 3579 & 17.28 & 6597  \\
11072 & 161.29 & 5663  & 24.39 & 5158  & 18.64 & 3119 & 17.26 & 5585  \\
12072 & 169.55 & 5567  & 26.70 & 5111  & 18.93 & 2920 & 17.24 & 5397  \\
13072 & 189.43 & 5044  & 29.33 & 4658  & 19.10 & 2735 & \textbf{17.15} & 4809  \\
14072 & 203.86 & 4915  & 32.21 & 4616  & OOM   & -    & 17.22 & 4866  \\
15072 & 221.14 & 4561  & 33.47 & 4292  & OOM   & -    & 17.23 & 4491  \\
16072 & 231.29 & 4382  & 34.51 & 4099  & OOM   & -    & 17.22 & 4312  \\

 \bottomrule
\end{tabular}
\end{center}
\end{table}

\clearpage 
\subsection{\al Results on WikiText-103}
\begin{figure}[h]
\begin{center}
\includegraphics[width=.5\textwidth]{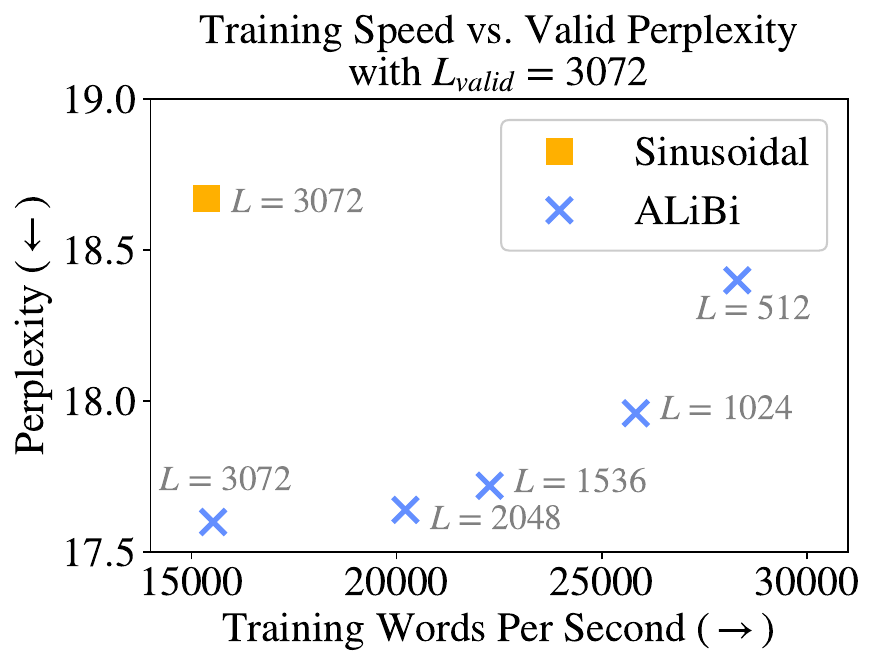}

\end{center}
\caption{The training speed and validation perplexity (with \li = 3072) for \al models and the sinusoidal model trained with \lt = 3072. All our models trained on 512 or more tokens achieve better perplexity than the sinusoidal model even though all of them (except the \lt = 3072) require less time and memory to train.  }
\label{fig:wt103_train_speed_vs_ppl}
\end{figure}

Figure~\ref{fig:wt103_train_speed_vs_ppl} depicts a cross section of Figure~\ref{fig:wt103_extra_lsb_all}, showing our models with different train lengths and the sinusoidal baseline, all evaluated on \li = 3072 tokens. We observe that all our models with 512 $\leq$  \lt $<$ 3072 are faster to train than the sinusoidal model with \lt = 3072, 
but they all achieve greater perplexity scores on the validation set. Our model with \lt = 3072 trains just as fast as the sinusoidal one but bests its score by more than one perplexity point; (the standard deviation for the the sinusoidal model with \lt = 3072 is 0.24).

Table~\ref{tab:LSB_wt103} shows the perplexity values obtained when 8 different \al models, trained on $L$ values between 64 and 3072, extrapolating to \li values longer than the ones they were trained on. In addition, we present results for the sinusoidal, rotary and T5 bias models, with \li = \lt. 
\begin{table}[h]
\caption{Perplexity when ALiBi extrapolates on the WikiText-103 development set. $^*$For results we present for the sinusoidal, rotary and T5 bias models, \lt = \li (so we do not test the extrapolation abilities of those baselines here). 
}
\label{tab:LSB_wt103}
\begin{center}
\begin{tabular}{@{}rcccccccc@{}}
\toprule
\multicolumn{1}{c}{ALiBi } &  \multicolumn{8}{c}{Evaluation Length} \\  
Train Length & 64 & 128 & 256 & 512 & 1024 & 1536 & 2048 & 3072 \\\cmidrule(r){1-1} \cmidrule(l){2-9}
64   &28.46&24.70&22.88&22.09&21.73&21.63&21.59&21.53\\
128  &-&23.98&21.70&20.67&20.36&20.29&20.31&20.28\\
256  &-&-&\textbf{21.29}&19.89&19.29&19.13&19.10&19.03\\
512  &-&-&-&19.73&18.81&18.50&18.48    &18.40\\
1024 &-&-&-&-&\textbf{18.66}&18.20&18.05        &17.96\\
1536 &-  &  -&-  &  -&-  &\textbf{18.12 }& 17.90&17.72  \\
2048 & - & - & - &  -& - & -&\textbf{17.91 }    &17.64 \\
3072 &  -&-  &  -&  -&  -&- &-         &\textbf{17.60} \\ 

\midrule %
\midrule
\multicolumn{1}{c}{Sinusoidal$^*$ }&\textbf{ 28.03}&\textbf{ 23.81}& 21.45& 20.05&19.34&19.05&18.87&18.67\\
\multicolumn{1}{c}{Rotary$^*$ } &- &-&-&20.07&19.33&-&-&18.57\\
\multicolumn{1}{c}{T5 Bias$^*$ }&- &-&-&\textbf{19.65}&18.80&-&-&18.01\\

\bottomrule
\end{tabular}
\end{center}
\end{table}

Table~\ref{tab:wt103_test} compares \al to the sinusoidal, rotary and T5 bias baselines on the test set of WikiText-103, and Table~\ref{tab:wt103_test_sota} compares \al to the current state of the art models on that test set.

\definecolor{mg}{gray}{0.5}

\begin{table}[t]

\caption{Test perplexity and runtime on WikiText-103 for two of our ALiBi models and models that use the sinusoidal, rotary and T5 bias methods. }
\label{tab:wt103_test}
\begin{center}
\centering

\begin{tabular}{@{}lccccc@{}} \toprule
\multirow{2}{*}[-2pt]{Model}  &\multirow{2}{*}[-2pt]{Param. $\downarrow$}&Train&\multicolumn{3}{c}{Inference  } \\  \cmidrule(lr){3-3} \cmidrule(lr){4-6}

 & &Speed$\uparrow$ & Speed $\uparrow$& Valid $\downarrow$ & Test $\downarrow$ \\ \midrule

Sinusoidal, $L$ = 3072 &\textbf{247M}  & 15.3k &\textbf{13.6k}&18.67&19.38 \\      
Rotary, $L$ = 3072     &\textbf{247M} &\textbf{11.5k} &12.2k &18.57 & 19.28 \\  
T5 Bias, $L$ = 3072    &\textbf{247M} &4.3k &7.3k&18.01&18.73 \\    

\midrule
 &&&&& \\
\addlinespace[-0.75em]

\multirow{2}{*}[3pt]{\rotatebox[origin=c]{90}{ALiBi}} $L$ = 512, $L_{valid}$ = 3072 &\textbf{247M}&28.3k&\textbf{13.6k}&18.40& 19.08\\

~~~~$L$ = 3072, $L_{valid}$ = 3072 &\textbf{247M}&15.5k&\textbf{13.6k}&\textbf{17.60}&\textbf{18.30} \\

\addlinespace[-0.75em]
 &&&&& \\
\bottomrule

\end{tabular}

\end{center}
\end{table}

\begin{table}[!htbp]

\caption{Valid and test perplexity scores on WikiText-103 for two of our ALiBi  models and models that use the sinusoidal, rotary and T5 bias methods with sliding window evaluation (\S\ref{sec:analysis} and S=512 following~\citep{baevski,khandelwal20generalization, shortformer}). The sinusoidal model presents our results from training and inference with the model of Baevski \& Auli. }

\label{tab:wt103_test_sota}
\begin{center}
\centering

\begin{tabular}{@{}lccc@{}} \toprule

{Model}  &{Param. $\downarrow$}& Valid $\downarrow$ & Test $\downarrow$ \\   \midrule
Adaptive Inputs~\citep{baevski} &\textbf{247M}   &17.97 &18.70\\
{Transformer-XL~\citep{transformer-xl}} &{257M }& - & 18.3\\
{Shortformer~\citep{shortformer}}&\textbf{247M}& 17.47&18.15  \\
{Sandwich Transformer~\citep{sandwich}}&\textbf{247M}&- &17.96 \\
Staged Training~\citep{shortformer}&\textbf{247M}&- &17.56 \\ 
{Compressive Transformer~\citep{Rae2020Compressive}}&329M         &-&17.1\\
Routing Transformer~\citep{roy2020efficient}& - &  - & 15.8\\
{kNN-LM~\citep{khandelwal20generalization}     }&\textbf{247M}&  15.81  & \textbf{15.79}\\ \midrule

Sinusoidal, $L$ = 3072 &\textbf{247M}  & 17.95&18.67 \\      
Rotary, $L$ = 3072     &\textbf{247M} & 17.98&18.72\\  
T5 Bias, $L$ = 3072    &\textbf{247M} &17.37& 18.12\\    

\midrule
 &&& \\
\addlinespace[-0.75em]

\multirow{2}{*}[3pt]{\rotatebox[origin=c]{90}{ALiBi}}
$L$ = 512, $L_{valid}$ = 3072 &\textbf{247M}&18.30&19.01\\

~~~~$L$ = 3072, $L_{valid}$ = 3072 &\textbf{247M}&16.97&17.66 \\

\addlinespace[-0.75em]
 &&& \\
\bottomrule

\end{tabular}

\end{center}
\end{table}

\subsection{Results on the Toronto Book Corpus}
\label{subsec:tbc}

To ensure that our results are not specific to the WikiText-103 corpus, we next apply our model and the baselines to a different domain while using a similar model architecture and the same ALiBi slopes as those used in the previous subsection.

We emphasize that our set of slopes was chosen by running experiments on the WikiText-103 corpus, and here we apply that set of slopes to a model trained on a very different text domain. Throughout the entire process of developing this method, we ran only one set of experiments on this domain using the previously selected set of slopes. %

Specifically, we use the Toronto BooksCorpus~\citep{zhu2015aligning}, which has been used to train BERT~\citep{bert} (in conjuction with the English Wikipedia). The corpus is about 700M tokens (2.9 GB).

We use the same train/validation/test split as~\citet{khandelwal20generalization} and their tokenization, which uses BERT's vocabulary of 29K byte-pair encodings.
Since the vocabulary is much smaller than WikiText-103's, we replace the adaptive word embedding and softmax of \citet{baevski} with a tied word embedding and softmax matrix \citep{tying,inan2017}.

\begin{figure}[h]
\begin{center}
\includegraphics[width=.75\textwidth]{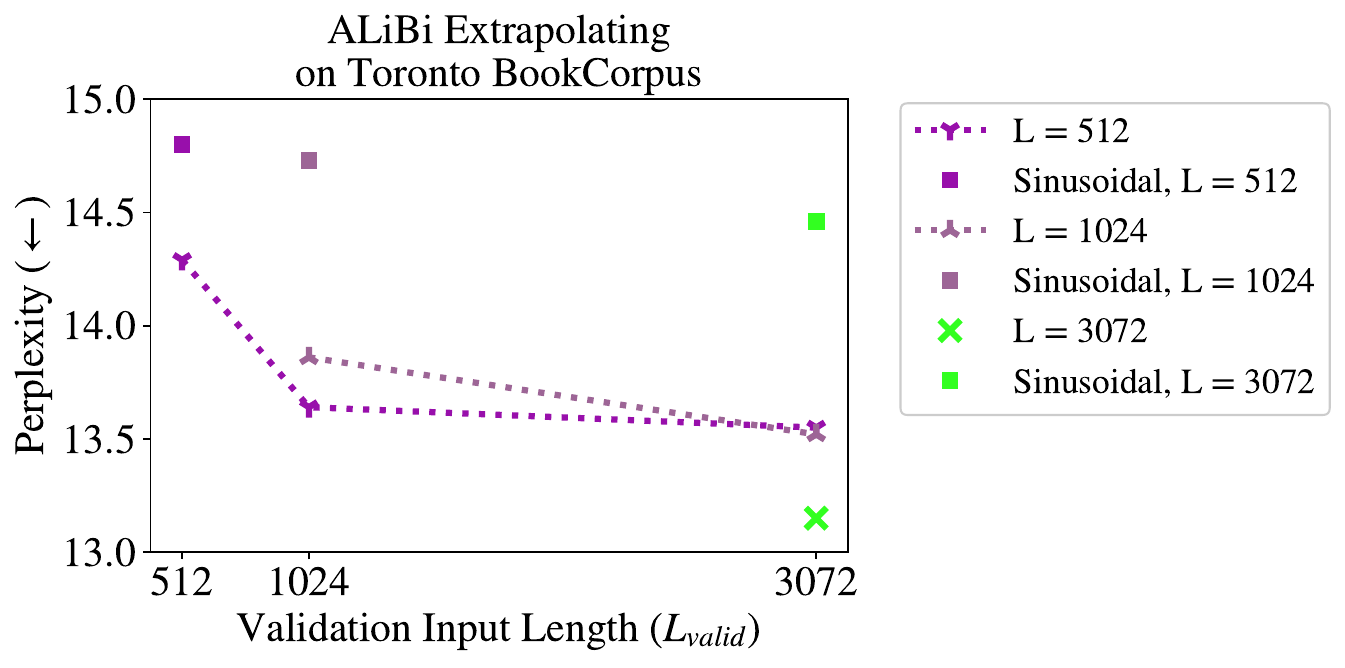}
\end{center}
\caption{ALiBi-enabled models evaluated on different input lengths on the Toronto BookCorpus. %
Our models extrapolate to longer sequence lengths and outperform the sinusoidal baseline even when trained on much shorter sequences. }
\label{fig:tbc_extra_lsb_all}
\end{figure}

Our results in Figure~\ref{fig:tbc_extra_lsb_all} (and Table~\ref{tab:books_lsb})  replicate our success on the WikiText-103 dataset. Our model surpasses the sinusoidal baseline when trained on the same amount of input tokens ($L$) and, in addition, our model is able to extrapolate to longer sequences at inference. This occurs even though our set of slopes was \emph{not} tuned on this dataset. %
This result establishes the generality of ALiBi and the particular set of slopes we found and suggests that they may be used on different text domains without further hyperparameter tuning.

Tables~\ref{tab:books_test} and \ref{tab:books_test_sota} present the perplexities for our \al models, the baselines, and the current state of the art on the Toronto BookCorpus validation and test sets. Our results here mirror our results on WikiText-103: we improve over the sinusoidal baseline even when AliBi is trained on fewer tokens. %

\begin{table}[!htbp]
\caption{\al   models extrapolating on the Toronto BookCorpus development set. $^*$For the results of the sinusoidal models, \lt = \li (so we do not test the extrapolation abilities of those models here).}
\label{tab:books_lsb}
\begin{center}
\begin{tabular}{@{}rccc@{}}
\toprule
                               &  \multicolumn{3}{c}{Evaluation Length} \\  
               Train Length               & 512 & 1024  & 3072 \\\cmidrule(r){1-1} \cmidrule(l){2-4}

512 & 14.29 &13.64 &13.55  \\
 1024 & -&13.86 & 13.52 \\
3072 & -& -& 13.15\\ \midrule \midrule
Sinusoidal$^*$ &14.80 &14.73 & 14.46 \\
\bottomrule
\end{tabular}
\end{center}
\end{table}

\begin{table}[!htbp]
\caption{Validation and test perplexities on the Toronto Book Corpus dataset. %
}
\label{tab:books_test}
\begin{center}
\centering

\begin{tabular}{@{}lccc@{}} \toprule
{Model}  &{Param. $\downarrow$}& Valid $\downarrow$ & Test $\downarrow$ \\ 

 \midrule
Sinusoidal, L = 3072 &\textbf{247M} & 14.46&11.67  \\

\midrule

 &&& \\
\addlinespace[-0.75em]

\multirow{2}{*}[3pt]{\rotatebox[origin=c]{90}{ALiBi}} $L_{train}$ = 512, $L_{valid}$ = 3072 &\textbf{247M}&13.55 & 10.98 \\
~~~~$L_{train}$ = 3072, $L_{valid}$ = 3072 &\textbf{247M}&\textbf{13.15} &\textbf{10.73} \\

\addlinespace[-0.75em]
 &&& \\
\bottomrule

\end{tabular}

\end{center}
\end{table}

\begin{table}[t]
\caption{Validation and test perplexities on the Toronto Book Corpus dataset with a sliding window (\S\ref{sec:analysis}). Following~\citep{baevski,khandelwal20generalization,sandwich,shortformer}, we set the sliding window stride $S$=512. %
}
\label{tab:books_test_sota}
\begin{center}
\centering

\begin{tabular}{@{}lccc@{}} \toprule
{Model}  &{Param. $\downarrow$}& Valid $\downarrow$ & Test $\downarrow$ \\ \midrule
kNN-LM~\citep{khandelwal20generalization} &\textbf{247M} &14.20&10.89 \\ 
Shortformer~\citep{shortformer} &\textbf{247M} &13.40&10.88 \\ 
Sandwich~\citep{sandwich} &\textbf{247M} &-&10.83 \\ 
Staged Training~\citep{shortformer}&\textbf{247M}  &12.80&10.48 \\

 \midrule
Sinusoidal, L = 3072 &\textbf{247M} & 14.06 & 11.40   \\

\midrule

 &&& \\
\addlinespace[-0.75em]

\multirow{2}{*}[3pt]{\rotatebox[origin=c]{90}{ALiBi}} 
$L$ = 512, $L_{valid}$ = 3072 &\textbf{247M}&13.76 & 11.11 \\
~~~~$L$ = 3072, $L_{valid}$ = 3072 &\textbf{247M}&12.70 & 10.40 \\

\addlinespace[-0.75em]
 &&& \\
\bottomrule

\end{tabular}

\end{center}
\end{table}

\begin{table}[!htbp]

\caption{Perplexity, memory, and train time on the CC100+RoBERTa corpus for our \al models and the sinusoidal baseline. We run our \lt = 512 (1024) model and the sinusoidal model with \lt = 1024 (2048) for the same amount of time. We show that our models achieve strong results even though they use 6--11\% less memory. }
\label{tab:ccr}
\begin{center}
\centering

\begin{tabular}{@{}lccccc@{}}
\toprule
&  \multicolumn{3}{c}{Training}&  \multicolumn{2}{c}{Valid PPL $\downarrow$}\\ \cmidrule(l){2-4} \cmidrule(l){5-6} 
&   { Memory $\downarrow$} & {Updates} &{ Hours $\downarrow$} &\li = 1024  & \li = 2048   \\             
\midrule
Sinusoidal, $L_{train}$ = 1024  &26.2 GB &46.7k &5.5k &  \textbf{9.24}  & -   \\              
ALiBi, $L_{train}$ = 512        &\textbf{24.6 GB}& 50.0k&5.5k&  9.30  & -   \\
 
\midrule
Sinusoidal, $L_{train}$ = 2048  &29.3 GB &44.2k &  5.9k&-&   9.01\\ 
ALiBi, $L_{train}$ = 1024        &\textbf{26.2 GB}&50.0k & 5.9k &-&  \textbf{8.92}  \\
\bottomrule
\end{tabular}

\end{center}
\end{table}

\subsection{Results on the CC100+RoBERTa Corpus}
Table~\ref{tab:ccr} compares our 1.3 billion parameter \al models when extrapolating to two times the number of tokens that they were trained on. We use the sinusoidal model as our baseline, and train it for the same amount of time as we train the \al model that we compare it to (and so since our \al models run faster in this setting, the sinusoidal models complete less updates). 

Table~\ref{tab:ccr_all_50k} compares our 1.3 billion parameter \al models to the sinusoidal baselines, with and without extrapolation, with all models completing 50,000 updates. 
\begin{table}[!htbp]

\caption{Perplexity, train time and memory use of the sinusoidal and ALiBi models on the CC100+RoBERTa corpus when all models are trained with 50k updates.}
\label{tab:ccr_all_50k}
\begin{center}
\centering
\small
\setlength{\tabcolsep}{2pt}

\begin{tabular}{@{}lcccccc@{}}
\toprule
&  \multicolumn{3}{c}{Training}&  \multicolumn{3}{c}{Valid PPL $\downarrow$}\\ \cmidrule(l){2-4} \cmidrule(l){5-7} 
&   { Memory $\downarrow$} & {Updates} &{ Hours $\downarrow$} &\li = 512  &\li = 1024  & \li = 2048   \\             
\midrule
Sinusoidal, $L_{train}$ = 512  &24.6 GB & 50.0k & 5.5k &9.71&37.05 &105.42   \\

ALiBi, $L_{train}$ = 512        &24.6 GB& 50.0k& 5.5k & 9.79 &9.30&9.54   \\

\midrule
Sinusoidal, $L_{train}$ = 1024  &  26.2 GB &50.0k & 5.9k&-&9.15  &48.85\\
ALiBi, $L_{train}$ = 1024        &  26.2 GB &50.0k & 5.9k &-&9.16&8.92  \\

\midrule
Sinusoidal, $L_{train}$ = 2048  & 29.3 GB &50.0k &6.7k  & -&-&8.83\\ 
ALiBi, $L_{train}$ = 2048        & 29.4 GB&50.0k & 6.7k  & -&-&8.84  \\  
\bottomrule
\end{tabular}

\end{center}
\end{table}

\section{Analysis}
\label{sec:analysis}

In this section we investigate why \al works so effectively.  
We find that ALiBi's decrease in perplexity when given longer sequences is largely explained by its improved avoidance of the early token curse.  We hypothesize that future work building on \al might achieve further gains by more efficiently  exploiting longer histories.

\subsection{Defining sliding window evaluation and the early token curse}

\begin{figure}[h]
\begin{center}
\includegraphics[width=.5\textwidth]{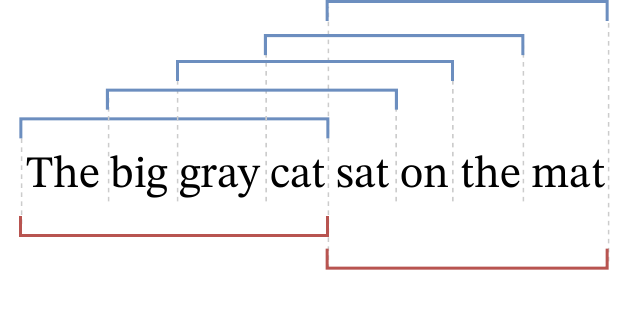}
\end{center}
\caption{Sliding window evaluation (top; blue) compared to nonoverlapping evaluation (bottom; red) on a sequence of 8 words using a  model with \li = 4. Nonoverlapping evaluation is much faster since it requires just two inference passes (as opposed to the five passes required by the siding window approach). But the sliding window approach provides more context for each prediction.}
\label{fig:bigcat}
\end{figure}

\paragraph{Sliding Window Inference}
As mentioned in Section~\ref{sec:act2}, nonoverlapping inference is commonly used to evaluate sequences longer than $L$ (the number of tokens in each training subsequence). 
An alternative is to use a sliding window during evaluation~\citep{baevski}. 

A stride $S$ is picked between 1 and $L-1$, and the window is advanced by $S$ tokens after each forward pass.\footnote{Nonoverlapping inference can be viewed as sliding window inference with stride $L$.} This means that $L-S$ tokens from the previous subsequence are re-encoded, and only $S$ new tokens are output. The advantage is that all outputs in each subsequence after the first have at least $L-S$ previous tokens to condition on. However, since tokens must be re-encoded multiple times, this approach is much slower than the nonoverlapping one. 
When $S=1$, we output one token every inference pass, each using the maximal context window that the model can handle; however, this is the slowest approach. 
Figure~\ref{fig:bigcat} is a visualization of the nonoverlapping and sliding window evaluation approaches. 

We use sliding window inference as a tool to analyze our models, but we note that it is normally prohibitively slow in practice~\citep{shortformer}. 

\paragraph{Early Token Curse}
Splitting an evaluation set into subsequences means that predictions occuring early in each subsequence cannot access many previous context tokens (appearing at the end of the previous subsequence). 
The result, referred to as the \emph{early token curse}~\citep{shortformer}, increases (i.e., degrades) perplexity scores. %
A workaround is to evaluate the model using a sliding window, giving each prediction more context. This solution is slow since it requires many more forward passes of the model.

\subsection{Extrapolation reduces the early token curse}

We presented results showing that our ALiBi method (and, to a lesser extent, the T5 bias) allows LMs to extrapolate during inference. Two reasons could explain why these methods enable LMs to achieve better perplexity given longer input subsequences:
\begin{enumerate}
  \item Performance improves because the models can use longer contexts to make more accurate predictions. For example, the average article length in the WikiText-103 corpus is about 3600 tokens; therefore, if a model trained on $L=512$ tokens extrapolates to \li = 3072 tokens during inference and achieves better results, that might be because it can spot patterns occurring across more than 512 tokens.
  \item Performance improves because longer input sequences mean the early token curse is reduced. %
  For example, during nonoverlapping evaluation on sequences of length \li $=$ 1000, 10\% of predictions have 100 tokens of context or less. If we rerun nonoverlapping evaluation on that model with \li $=$ 2000 tokens, now only 5\% of predictions have 100 tokens of context or less. So, by simply being able to handle longer sequences, a model can substantially reduce the early token curse and improve performance.\footnote{100 tokens is an arbitrary small number used here to represent a short history context, i.e., one in which making predictions for the next output token would be harder.}
\end{enumerate}

To better understand what might be occurring, we re-evaluate the development set of WikiText-103 with our models and the sinusoidal baseline with $L = 512, 1024, 3072$. However, this time we use sliding window evaluation with a stride of $S=1$, meaning that we move the sliding window just one token after every inference pass, giving each prediction the maximum number of context tokens that the model can use. 

\begin{figure}[h]
\begin{center}
\includegraphics[width=.75\textwidth]{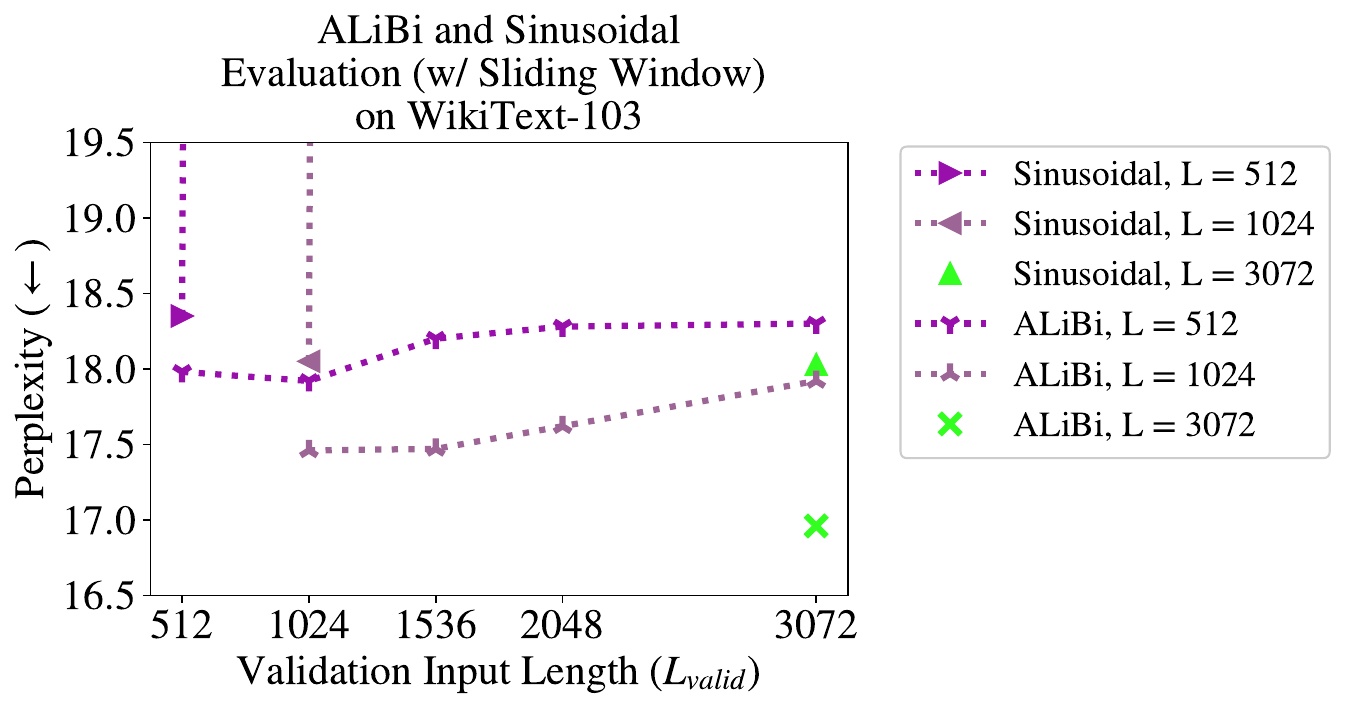}

\end{center}
\caption{ALiBi models evaluated on different input lengths on WikiText-103 with sliding window evaluation (with stride $S=1$). Unlike results shown in Figure~\ref{fig:wt103_extra_lsb_all}, where performance improves in each of our models as we increase the validation sequence length, here performance stays relatively flat as we increase \li. This might mean that \al increases performance when \li $>$ \lt not because it uses longer contexts, but because  %
fewer tokens suffer from the early token curse. Note that as in \S\ref{sec:act2}, the perplexity of the sinusoidal model explodes when \li $>$ \lt even when using sliding window evaluation.}
\label{fig:wt103_extra_lsb_cw}
\end{figure}

The results are shown in Figure~\ref{fig:wt103_extra_lsb_cw} and in the corresponding Tables~\ref{tab:sin_extrapolating_cw} (sinusoidal) and \ref{tab:lab_extrapolating_cw} (ALiBi). 

Unsurprisingly, for the sinusoidal model, as in \S\ref{sec:act2}, increasing \li causes an explosion in perplexity even when using sliding window evaluation.
Our ALiBi models cannot improve perplexity when looking at longer sequences in this setting, but they keep perplexity flat when \li increases.

This leads us to believe that our perplexity improvement when increasing \li and using nonoverlapping evaluation is caused by explanation 2, not explanation 1. Because sliding window evaluation provides long context windows for \emph{every} prediction made, it curtails the early token curse. In this setting, \al's performance remains flat when \li increases, leading us to hypothesize that the gains seen while increasing \li in \S\ref{sec:results} were the result of larger \li values mitigating the early token curse. 

Our \al results mirror what occurs in the model using the T5 bias: when using sliding window evaluation, perplexity remains relatively flat when evaluating longer sequences (see Table~\ref{tab:t5_extrapolating_cw}).

Our analysis reveals that when $\li > \lt$, \al might not be using contexts longer than the ones it was trained on. This highlights a research direction that could be pursued in future work.

These findings do not lessen the value of \al.  When \li = \lt, \al achieves either superior or similar results to the sinusoidal method and other alternatives even though it is simpler and requires no learned parameters.
When evaluating $\li > \lt$ tokens, even if \al does not attend to more than \lt tokens, it yields better results than the other alternatives that can be used in this case, i.e., standard nonoverlapping inference (which is cheap, but does not perform as well) and the more accurate sliding window approach (which is very slow).

\begin{table}[t]
\caption{Perplexities of the \textbf{sinusoidal} models evaluated with sliding window evaluation with stride $S=1$ on the WikiText-103 validation dataset.}
\label{tab:sin_extrapolating_cw}
\begin{center}
\begin{tabular}{@{}rccccc@{}}
\toprule
                               &  \multicolumn{5}{c}{Evaluation Length ($S=1$)} \\  
               Train Length               & 512 & 1024 & 1536 & 2048 & 3072 \\\cmidrule(r){1-1} \cmidrule(l){2-6}

512& 18.35& 204.42& 264.74& 306.19& 360.12\\
 1024 &-& 18.05& 206.55& 302.6& 393.71\\
3072 & -&- & -& -& 18.03 \\ \bottomrule
\end{tabular}
\end{center}
\end{table}

\begin{table}[t]
\caption{Perplexities of the \textbf{T5 bias} models evaluated with sliding window evaluation with stride $S=1$ on the WikiText-103 validation dataset. }
\label{tab:t5_extrapolating_cw}
\begin{center}

\begin{tabular}{@{}rccccc@{}}
\toprule
                               &  \multicolumn{5}{c}{Evaluation Length ($S=1$)} \\  
               Train Length               & 512 & 1024 & 1536 & 2048 & 3072 \\\cmidrule(r){1-1} \cmidrule(l){2-6}

512&  17.92 & 18.51 & 20.36 & 22.62 & 30.77 \\
 1024 &-& 17.65 & 17.87 & 18.51 & 20.66  \\
3072 & -&- & -& -&  17.41\\ \bottomrule
\end{tabular}

\end{center}
\end{table}

\begin{table}[t]
\caption{Perplexities of the \textbf{ALiBi} models evaluated with sliding window evaluation with stride $S=1$ on the WikiText-103 validation dataset.}
\label{tab:lab_extrapolating_cw}
\begin{center}
\begin{tabular}{@{}rccccc@{}}
\toprule
                               &  \multicolumn{5}{c}{Evaluation Length ($S=1$)} \\  
               Train Length               & 512 & 1024 & 1536 & 2048 & 3072 \\\cmidrule(r){1-1} \cmidrule(l){2-6}

512& 17.98 & 17.92 & 18.2 & 18.28 & 18.3 \\
 1024 &-& 17.46 & 17.47 & 17.62 & 17.92  \\
3072 & -&- & -& -& 16.96\\ \bottomrule
\end{tabular}
\end{center}
\end{table}

\end{document}